\documentclass{article} 
\usepackage{iclr2027_conference,times}

\usepackage{amsmath,amsfonts,bm}

\def\eqref#1{equation~\ref{#1}}

\def\1{\bm{1}}

\DeclareMathAlphabet{\mathsfit}{\encodingdefault}{\sfdefault}{m}{sl}
\SetMathAlphabet{\mathsfit}{bold}{\encodingdefault}{\sfdefault}{bx}{n}

\usepackage{url}            
\usepackage{booktabs}       
\usepackage{multirow}
\usepackage{amsfonts}       
\usepackage{nicefrac}       
\usepackage{microtype}      
\usepackage{xcolor}         
\usepackage{graphicx}
\usepackage{amsmath}
\usepackage{amssymb}
\usepackage{amsthm}
\usepackage{mathtools}
\usepackage{etoc}
\usepackage{tabularx}
\usepackage{algorithm}
\usepackage{enumitem}
\usepackage[noend]{algpseudocode}

\usepackage[pagebackref=true,breaklinks=true,colorlinks=true,bookmarks=false]{hyperref}
\definecolor{deepred}{HTML}{940000}
\hypersetup{linkcolor=deepred}
\hypersetup{urlcolor  = [rgb]{0.4,0.15,0.95}}
\hypersetup{citecolor=[rgb]{0.4,0.15,0.95}}
\usepackage[capitalize,noabbrev]{cleveref}

\usepackage{pifont}
\usepackage{makecell}
\usepackage{titletoc}

\usepackage[toc,page,header]{appendix}
\usepackage[nohints]{minitoc}

\renewcommand \thepart{}
\renewcommand \partname{}

\usepackage{caption}

\definecolor{Gray}{gray}{0.92}
\newlength\savewidth\newcommand\shline{\noalign{\global\savewidth\arrayrulewidth
  \global\arrayrulewidth 1pt}\hline\noalign{\global\arrayrulewidth\savewidth}}

\title{\fontsize{17pt}{\baselineskip}\selectfont What Makes Recurrence Effective in Looped\\Language Models?}

\newcommand\blfootnote[1]{%
  \begingroup
  \renewcommand\thefootnote{}\footnote{#1}%
  \addtocounter{footnote}{-1}%
  \endgroup
}

\iclrfinalcopy 

\author{ \fontsize{9.5pt}{\baselineskip}\selectfont
    Xinlin Zhuang\textsuperscript{1,2} \quad
    Siyuan Wang\textsuperscript{1} \quad 
    Imran Razzak\textsuperscript{2} \quad
    Weiyang Liu\textsuperscript{1,*} \\[0.5mm]
    \fontsize{9.5pt}{\baselineskip}\selectfont
    \textsuperscript{1}The Chinese University of Hong Kong \quad
    \textsuperscript{2}MBZUAI
}

\begin{document}

\maketitle
\doparttoc
\faketableofcontents
\lhead{~\raisebox{-0.47ex}{\includegraphics[height=.38cm]{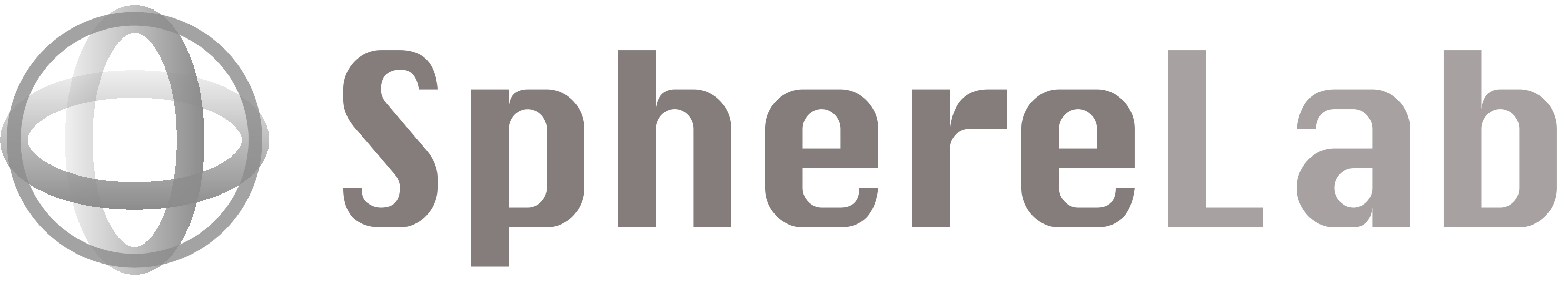}}~~~Technical Report}    

\blfootnote{\textsuperscript{*}Corresponding author \quad Project page:~{\tt\href{https://spherelab.ai/looplm}{\textbf{spherelab.ai/looplm}}}}


\vspace{-8mm}
\begin{abstract}
\vspace{-0.6mm}
Looped language models (LoopLMs) increase computational depth through parameter sharing, offering a path to scale inference computation without adding parameters.
However, it remains unclear when additional recurrence is beneficial and how architectural choices affect its effectiveness.
Through controlled experiments, we systematically examine (1) when recurrence helps, (2) where it should be applied, and (3) how its conditioning affects performance.
Our evaluation covers inference budgets below, within, and beyond the training horizon under knowledge and reasoning tasks.
(1) We find that recurrence can improve reasoning beyond the training horizon while degrading knowledge performance, but harder reasoning instances do not consistently benefit more. 
(2) Performance also depends on how distinct layers and recurrent iterations are allocated, showing that effective depth alone is insufficient to predict behavior.
Non-recurrent output layers improve robustness to under-unrolling, while the preferred placement of input and output layers varies with inference budget.
(3) Finally, we find that conventional initial-state injection offers limited robustness to varying recurrence depth. 
We therefore propose history-state injection as an alternative, and show that channel-wise history-state injection combined with timestep conditioning offers a low-cost and more effective design, better preserving knowledge under extended unrolling while improving robustness across inference budgets.
Overall, our results clarify when recurrent computation helps, where it fails, and offer practical guidelines for designing LoopLMs across variable inference budgets.
\end{abstract}

\vspace{-1.5mm}
\section{Introduction}
\vspace{-.75mm}
Scaling large language models (LLMs) has driven substantial capability gains, but at the cost of rapidly growing parameter and memory requirements, making further scaling increasingly costly to train and deploy. 
Looped language models (LoopLMs) have emerged as a parameter-efficient alternative by repeatedly executing a shared stack of Transformer blocks~\citep{huginn,baseloop,ouro,loopformer}. 
By reusing parameters across iterations, LoopLMs enable deeper computation with a smaller memory footprint. 
More importantly, their recurrent structure allows inference depth to be flexibly adjusted by varying the loop count, providing a natural mechanism for scaling test-time computation~\citep{rins}.

However, the potential of LoopLMs for test-time scaling remains underexplored.
Existing work primarily evaluates models at the fixed recurrent depth during training (hereafter referred to as the \textit{training horizon}) or focuses on early exiting, leaving it unclear whether additional depth beyond the training horizon continue to provide useful computation and performance gains, particularly for computation-intensive tasks. 
Meanwhile, recent LoopLM architectures explores a growing design space, varying the size and count of recurrent blocks, where recurrence is placed within the network, and how recurrent computation is conditioned across iterations. 
While these designs can improve performance within the training horizon, they do not necessarily guarantee improvement during continued test-time unrolling. 
This raises a fundamental question: 
\textit{can recurrence remain effective as inference computation scales beyond its training horizon, and what factors determine this behavior?}

We begin by characterizing test-time recurrent scaling using a minimalist LoopLM without specialized architectural designs (Sec.~\ref{sec:motivation}).
We pre-train variants configured with different combinations of physical depth (the number of distinct layers) and loop count under a fixed effective training depth (the product of physical depth and loop count), and vary their inference depth.
Surprisingly, even this simple design scales substantially beyond its training horizon, with continued test-time unrolling further improving performance and certain configurations even outperforming non-recurrent counterparts under equivalent training compute.
However, this scaling behavior is not universal.
Reasoning tasks can benefit from additional loops while knowledge-oriented tasks degrade, and greater reasoning depth does not consistently lead to larger gains from further unrolling.
Moreover, different combinations of physical depth and loop count exhibit markedly different scaling behaviors despite sharing the same effective training depth.
These observations reveal both the potential and instability of recurrent test-time scaling: additional loops can unlock extra computation, their efficacy is strongly conditioned on task semantics and how recurrence is configured.

Motivated by this variability, we systematically investigate two factors that shape test-time recurrent scaling (Sec.~\ref{sec:comparison},~\ref{sec:features}):
\textit{which computations should be recurrent}, and \textit{how recurrent computation should be conditioned}.
For the former, we find that independently parameterized blocks surrounding the recurrent core play distinct roles across inference regimes.
Output-side non-recurrent blocks improve knowledge robustness to under-unrolling, whereas allocating more non-recurrent computation to the input side tend to better support reasoning extrapolation; near the training horizon, performance is less sensitive to their allocation.
For the latter, we observe that initial-state injection degrades knowledge robustness during loop extrapolation when over-parameterized. 
We therefore propose \textbf{history-state injection}, which conditions on relative differences from intermediate states to capture dynamic trajectories and substantially rescues deep extrapolation performance. Timestep conditioning further improves reasoning extrapolation by allowing the shared computation to vary across recurrent iterations, although its effectiveness depends on the recurrent configuration.

Building on these insights, we introduce a lightweight conditioning framework that jointly incorporates history-state injection and timestep conditioning through channel-wise parameterization. 
This design alleviates the limitations of either conditioning scheme alone, yielding complementary gains and consistently outperforming both individual variants and the unconditioned BaseLoop baseline under extended inference budgets (Fig.~\ref{fig:combine_results}(\textbf{c})). 
In summary, our contributions are threefold:
\begin{itemize}[leftmargin=*,itemsep=2pt,topsep=2pt]
    \item \textbf{Characterizing recurrent scaling.} We present a systematic empirical study characterizing LoopLMs beyond their training horizon, showing substantial test-time scaling potential but also strong dependence on specific tasks, reasoning depth, and recurrent configuration.
    \item \textbf{Understanding effective recurrence.} We identify key factors, including non-recurrent block allocation, dynamic state history, and timestep conditioning, that govern extrapolation behavior across knowledge and reasoning tasks.
    \item \textbf{Designing a synergistic LoopLM.} We introduce a lightweight joint conditioning mechanism combining our proposed history-state injection and timestep conditioning, achieving robust performance gains on both knowledge and reasoning tasks across varied inference budgets.
\end{itemize}
\vspace{-1mm}
\section{Preliminaries and Evaluation Settings}
\label{sec:preliminaries}
\vspace{-.75mm}
\subsection{Formulation}
\label{sec:formulation}
\vspace{-.5mm}

Following~\citet{huginn}, a LoopLM causal decoder typically comprises three groups of Transformer blocks: a $p$-block \textit{Prelude} $P$, an $s$-block parameter-shared \textit{recurrent core} $F$, and a $c$-block \textit{Coda} $C$. 
Omitting the token embedding $e(\cdot)$, the final normalization, and the language-model head $W$, which are identical across all models we study, a looped decoder can be written as $\mathcal{M}_{p,s,c,K} \;=\; C \circ F^{K} \circ P$. 
Here, $F^{K}$ denotes $K$ successive applications of the shared core and $K$ is the loop count during training, termed the \textit{training horizon}. 
At inference, the model can execute a varying number of loops $r$, corresponding to \textit{under-unrolling} ($r<K$), evaluation at the \textit{training horizon} ($r=K$), or \textit{loop extrapolation} ($r>K$).
Given an input sequence $x$, the Prelude initializes the recurrent state as $\tilde{x}=P(e(x))\in\mathbb{R}^{n\times d}$. 
The recurrent computation then evolves as
\begin{equation}
  h^{0} = \tilde{x};\qquad
  h^{t+1} = F\!\left(h^{t};\,\gamma_t\right),\quad 0\le t < r;\qquad
  y = W\, C(h^{r});
  \label{eq:loop-state-evolution}
\end{equation}
where $\gamma_t$ denotes auxiliary signals that condition the recurrent trajectory, such as the initial state, past recurrent states, or timestep information~\citep{huginn,attractor,expressive_power}. 
$\gamma_t$ is optional and setting $\gamma_t=\varnothing$ recovers a standard recurrence $h^{t+1}=F(h^{t})$. 
We discuss different forms of recurrent conditioning strategies and their impacts in Sec.~\ref{sec:features}.
 
We distinguish the model's \textit{physical} depth, $L_{\mathrm{phys}} = p+s+c,$ from its \textit{effective} depth, $L(r) = p + r \cdot s + c $. 
The former determines the number of independently parameterized Transformer layers, while the latter characterizes the real per-token compute.
Accordingly, $L(K)$ denotes the training effective depth and varying $r$ controls the inference effective depth $L(r)$.
A special case of LoopLM is when $p=c=0$, i.e.\ $\mathcal{M}_{0,s,0,K}=F^{K}$~\citep{baseloop}, where the entire network acts as the shared recurrent core, with physical depth $s$, effective depth $K \cdot s$, and no independently parameterized blocks surrounding the recurrence. 
We term this configuration \textit{BaseLoop} and refer to LoopLMs with $p+c>0$, which places non-recurrent boundary blocks on one or both sides of the core, as \textit{CoreLoop} for later comparisons.
Detailed comparisons between them are discussed in Sec.~\ref{sec:comparison}.

\vspace{-.5mm} 
\subsection{Controlled Experimental Setup}
\label{sec:setup}
\vspace{-.5mm} 
 
As recurrence trades parameters against compute, comparing architectural variants is meaningful only under controlled computation budgets.
Every comparison in this paper matches four quantities:
(i) physical depth $L_{\mathrm{phys}}=p+s+c$;
(ii) training effective depth $L(K)=p+K\cdot s+c$; 
(iii) inference effective depth $L(r)$; 
and (iv) the training pipeline, including the token budget, optimizer, etc. 
Further details of model architectures, data composition, training and evaluation are provided in App.~\ref{app:exp_settings}.

\textbf{Models and Data.}
We study LoopLMs in a controlled pre-training-from-scratch setup built on two dense backbone families: Llama3.1-1B~\citep{llama3} and Qwen3-0.6B~\citep{qwen3}. 
We adopt their layer configurations, widths, and tokenizers, while randomly initializing all parameters.
Models are trained on FineWeb-Edu~\citep{fineweb} with the standard next-token prediction objective, packing documents into 2048-token sequences without padding. 
Unless otherwise specified, the training token budget follows the Chinchilla ratio of 20 tokens per parameter, with parameter counts estimated from the model's effective training depth $L(K)$.
 
\textbf{Training.}
Each model is trained with a fixed loop count $K$, executing exactly $K$ recurrent iterations per sample without adaptive halting or early exits~\citep{mor}. 
Gradients are backpropagated through all $K$ iterations without truncation, and the causal language modeling loss is applied. 
We use the Muon optimizer~\citep{muon,dmuon} for hidden matrix parameters, and AdamW for embeddings, output heads, biases, and other non-matrix parameters.
All other optimization settings, including the learning-rate scheduler, warmup, and weight decay are held consistent for fair comparison. 
A detailed optimizer ablation for LoopLM training is provided in App.~\ref{app:muon}.

\begin{figure}[t!]
    \centering
    \setlength{\abovecaptionskip}{4pt}
    \setlength{\belowcaptionskip}{-4pt}
    \vspace{-1mm}
    \includegraphics[width=1.0\linewidth]
        {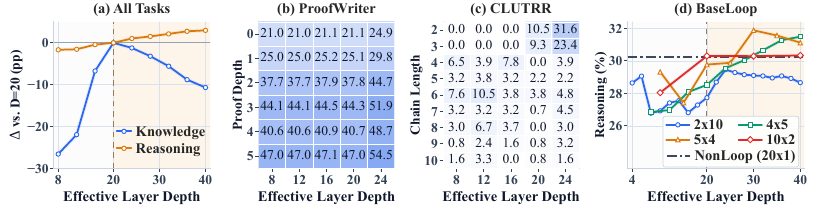}
    \caption{\footnotesize 
    Task- and architecture-dependent benefits of recurrence.
    \textbf{(a)} Score changes relative to $L(r)=20$ for BaseLoop $4\times5$ across knowledge and reasoning splits.
    \textbf{(b,c)} Absolute accuracy (\%) for
    BaseLoop $2\times10$ across inference depths, stratified by ProofWriter proof depth
    and CLUTRR chain length.
    \textbf{(d)} Reasoning scores across varying configurations of physical depth and
    loop count.
    In \textbf{(a,d)}, dashed vertical lines mark the training
    depth, and shaded parts indicate the depth extrapolation.
    }
    \label{fig:motivation}
\end{figure}

\vspace{-.5mm} 
\subsection{Evaluation across Computational Demands}
\vspace{-.5mm} 

To investigate whether and when additional recurrent computation benefits distinct task capabilities, we evaluate LoopLMs along two complementary axes: the \textit{computational demand of the task} and the \textit{executed depth at inference}. 
This reveals capability-specific dynamics that are otherwise obscured by the aggregate metrics in prior LoopLM evaluations~\citep{loopformer,huginn}.

\textbf{Knowledge vs. Reasoning.}
We partition downstream benchmarks into knowledge and reasoning groups according to the computation required to produce an answer. Knowledge tasks primarily rely on facts stored within model weights and require limited multi-step composition, whereas Reasoning tasks require composing information through multiple inference steps and may therefore benefit more from additional recurrent computation.
The knowledge group includes SciQ~\citep{sciq}, ARC-Easy~\citep{arc}, and PIQA~\citep{piqa}; the Reasoning group includes ARC-Challenge~\citep{arc}, WinoGrande~\citep{winogrande}, OpenBookQA~\citep{openbookqa}, HellaSwag~\citep{hellaswag}, CommonsenseQA~\citep{commonsenseqa}, ProofWriter~\citep{proofwriter}, CLUTRR~\citep{clutrr}, and BBH~\citep{bbh}.
We report the unweighted mean within each group, with the aggregate across groups as a summary.
Beyond this binary split, ProofWriter and CLUTRR provide controlled reasoning-depth axes, defined by proof depth and relation-chain length, respectively.
This allows us to examine whether the utility of additional recurrent computation changes with the amount of reasoning required by the task.

\textbf{Different Inference Budgets: Under-Unrolling, Training Horizon, and Loop Extrapolation.}
For each model trained with a fixed loop count $K$, we vary the inference loop count $r$ at test time without any additional training to evaluate three computational regimes: \textit{under-unrolling} ($r<K$), the \textit{training horizon} ($r=K$), and \textit{loop extrapolation} ($r>K$). 
We report performance as a function of the resulting effective depth $L(r)$, allowing us to track how different capabilities respond as recurrent computation is reduced, matched to training, or extended beyond the training horizon.
\vspace{-1mm}
\section{When Does Recurrence Help?}
\label{sec:motivation}
\vspace{-1mm}

To investigate whether and under what conditions additional recurrent depth provides useful test-time compute, we begin by analyzing BaseLoop models following the Llama3.1-1B architecture.
We train several variants with different loop configurations,
$K\times s\in
\{2\times10,4\times5,5\times4,10\times2\}$,
while fixing the effective training depth at
$L(K)=20$, where $K \times s$ denotes $s$ recurrent unrolls over a physical core of depth $K$.
All models share identical training configurations and are evaluated across under-unrolling, training horizon, and extrapolation settings.
A standard non-recurrent baseline ($20 \times 1$, termed \textit{NonLoop}) whose physical depth matches the effective depth of the loop models ($L(K) = 20$) serves as an upper-bound performance reference under equal compute.

\textbf{Additional recurrence can improve reasoning beyond
the training horizon.}
Fig.~\ref{fig:motivation}(\textbf{a}) illustrates how inference depth scaling affects knowledge and reasoning performance for BaseLoop $2\times10$.
Within the training horizon ($L(r) \le 20, r \le 4$), increasing inference depth
improves both task groups.
During extrapolation ($L(r) > 20, r > 4$), however, the two groups diverge.
Reasoning performance continues to improve, rising from 28.52 at $L(r) = 20$ to 31.49 at $L(r) = 40$, a 2.97 percentage point gain achieved purely at test-time without parameter updates. 
In contrast, knowledge performance decreases, dropping from 62.80 to 52.11.
Thus, the training horizon does not impose a strict ceiling on effective recurrence and test-time depth extrapolation selectively benefits tasks with higher computational demands while failing to scale static knowledge memorization.

\textbf{Gains vary across reasoning demands and complexities.}
The stratified results in Fig.~\ref{fig:motivation}(\textbf{b,c}) reveal that the additional recurrence also varies across reasoning benchmarks and complexity levels.
On ProofWriter, extending to the extrapolation range $L(r)=24$ consistently boosts accuracy across all proof depths (0–5), yielding absolute gains of around 7.5 percentage points at deeper proof depths (depth 3-5).
Conversely, the gains on CLUTRR vary significantly across relation-chain lengths.
Extrapolating to $L(r)=24$ yields large accuracy jumps on short chains, reaching 31.6 for length 2 and 23.4 for length 3, whereas longer chains exhibit substantially smaller gains or a decline (depths 4-10).
This shows that while extra recurrence helps reasoning, harder instances with long relation chains do not automatically benefit as much from simply adding inference loops.

\textbf{Physical depth and recurrent loops require a balanced allocation.}
Models trained at the same effective training depth ($L(K)=20$) exhibit markedly different scaling behaviors depending on how depth is allocated between physical layers and recurrent iterations (Fig.~\ref{fig:motivation}\textbf{d}).
BaseLoop $2\times10$, with a shallow physical core and many recurrent iterations, peaks early and degrades under further unrolling, whereas $10\times2$, with a deep physical core but few recurrent iterations, remains relatively flat during extrapolation, gaining little from additional loops.
More balanced configurations exhibit stronger test-time scaling: $5\times4$ peaks at $L(r)=30$, while $4\times5$ continues improving up to $L(r)=40$.
These results suggest that effective recurrent scaling requires a balanced allocation between physical depth and recurrent iterations, rather than being determined by effective depth alone.

\textbf{Additional computation enables recurrent models to surpass the non-recurrent upper-bound.}
With extra test-time compute, several recurrent configurations exceed the NonLoop reasoning score of 30.25 (Fig.~\ref{fig:motivation}\textbf{d}).
Specifically, BaseLoop $5\times4$ reaches 31.88 at $L(r)=30$, while $4\times5$ achieves 31.49 at $L(r)=40$.
Crucially, these recurrent variants are trained under the exact same effective depth and training FLOPs ($L(K)=20$) as NonLoop ($20\times1$), yet rely on significantly fewer distinct physical Transformer layers. 
This comparison demonstrates that parameter-efficient recurrent models can effectively trade additional test-time computation for superior reasoning performance.
\begin{figure}[t!]
    \centering
    \setlength{\abovecaptionskip}{3pt}
    \setlength{\belowcaptionskip}{-1pt}
    \vspace{-1.5mm}
    \includegraphics[width=1.0\linewidth]{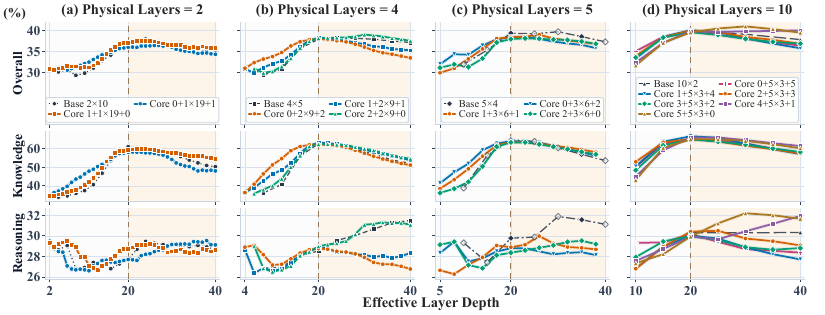}
    \caption{\footnotesize BaseLoop and CoreLoop performance based on the Llama3.1-1B architecture configuration on \textit{Overall}, \textit{Knowledge}, and \textit{Reasoning} benchmarks across inference depths. Dashed vertical lines mark the effective training depth $L(K)=20$, and shaded regions denote loop extrapolation.}
    \label{fig:baseloop_coreloop_main}
\end{figure}

\vspace{-1.5mm}
\section{Which Computations Should Be Recurrent?}
\label{sec:comparison}
\vspace{-1.5mm}

We next investigate whether all Transformer blocks should participate in recurrent weight sharing, or whether some computations are better implemented by independently parameterized boundary layers.
To this end, we compare \textit{BaseLoop}, which recurrently applies the entire Transformer stack, with \textit{CoreLoop}, which reserves non-recurrent Prelude and/or Coda blocks around the shared core.
Under matched effective training depth and token budget, we evaluate Llama3.1-1B and Qwen3-0.6B architecture configurations.
We present the Llama3.1-1B results in this section, considering physical depths of $\{2,4,5,10\}$ with the effective training depth fixed at $L(K)=20$, while varying the Prelude and Coda allocation.
Results for Qwen3-0.6B are provided in App.~\ref{app:baseloop_coreloop_results}.

At inference, we vary the loop count $r$ and evaluate knowledge, reasoning, and overall performance across under-unrolling, training horizon, and loop extrapolation. 
We further characterize recurrent representation dynamics using geometry metrics, including \textit{Angular Distance}, \textit{Relative Update Norm}, and \textit{Normalized State Variance}, as defined in App.~\ref{app:probing}.
Full results and comparisons are in App.~\ref{app:baseloop_coreloop_results}.

\begin{figure}[t!]
    \centering
    \setlength{\abovecaptionskip}{4pt}
    \setlength{\belowcaptionskip}{-6pt}
    \includegraphics[width=1.0\linewidth]{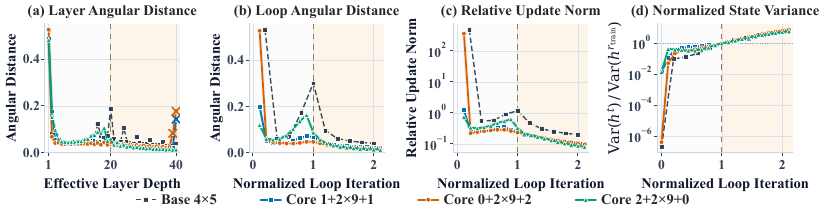}
    \caption{\footnotesize Geometric dynamics of BaseLoop and CoreLoop models with 4 physical layers from the Llama3.1-1B architecture.
    The horizontal axes in \textbf{(b)}-\textbf{(d)} are normalized by each configuration's training iterations.
\textbf{(a)} Angular distance between consecutive layer states, with crosses marking Coda layers.
\textbf{(b)} Angular distance between recurrent states before and after each complete loop iteration.
\textbf{(c)} Per-loop update magnitude relative to the preceding state.
\textbf{(d)} Recurrent-state variance at the training horizon. \textbf{(c-d)} use logarithmic scales.}
    \label{fig:baseloop_coreloop_geometry}
\end{figure}

As shown in Fig.~\ref{fig:baseloop_coreloop_main}, the optimal allocation of non-recurrent boundary layers varies with the inference budget. 
Across physical depths, three compute regimes (under-unrolling, the training horizon, and loop extrapolation) exhibit distinct trade-offs closely linked to representation dynamics (Fig.~\ref{fig:baseloop_coreloop_geometry}).
These metrics capture \textit{how much} recurrent states change, but not \textit{how strongly} later computation depends on earlier computation, a distinction we examine in Sec.~\ref{sec:mechanism}.

\textbf{Coda alleviates knowledge decay during under-unrolling.} 
BaseLoop's performance degrades rapidly with fewer inference loops, especially on knowledge tasks. 
Allocating non-recurrent layers to the Coda (e.g., $0+2\times9+2$ in Fig.~\ref{fig:baseloop_coreloop_main}\textbf{(b)}, $ 0+3\times6+2$ in Fig.~\ref{fig:baseloop_coreloop_main}\textbf{(c)}) substantially mitigates this degradation. 
Geometrically, CoreLoop configurations with a Coda block exhibit smaller angular changes and more stable recurrent states during under-unrolling than BaseLoop and the Prelude-only variant, as shown in Fig.~\ref{fig:baseloop_coreloop_geometry}\textbf{(a,b,c)}.
This suggests that separating the output-side transformation from the recurrent core helps align under-executed recurrent states with the final readout.

\textbf{Performance is strong and robust to boundary allocation at the training horizon.}
Around the effective training depth $L(K)=20$, models generally achieve strong and stable performance across overall, knowledge, and reasoning tasks, while different Prelude-Coda allocations become substantially smaller (Fig.~\ref{fig:baseloop_coreloop_main}).
This is the computation regime directly encountered during training, where the recurrent core produces representations well aligned with the trained output pathway. 

\textbf{Prelude-heavy allocations better support reasoning extrapolation.}
Beyond the training horizon, knowledge and overall performance generally deteriorate, whereas reasoning exhibits stronger and more configuration-dependent scaling.
Notably, allocating more non-recurrent capacity to the Prelude can better sustain or further improve reasoning performance under extended unrolling.
For example, the Prelude-heavy configuration $2+2\times9+0$ continues to improve in the four-layer setting, and $5+5\times3+0$ and $4+5\times3+1$ show strong reasoning extrapolation in the ten-layer setting (Fig.~\ref{fig:baseloop_coreloop_main}\textbf{(b,d)}).
Although not universal, this trend suggests that dedicated input transformations can improve recurrent computation beyond the training horizon.

\textbf{Convergence alone does not explain useful extrapolation.}
The representation dynamics reveal a notable discrepancy.
Beyond the training horizon, both loop angular distance and relative update norm progressively decrease (Fig.~\ref{fig:baseloop_coreloop_geometry}\textbf{(b,c)}), indicating increasingly small changes between recurrent iterations.
Meanwhile, recurrent-state variance continues to grow relative to its value at the training horizon (Fig.~\ref{fig:baseloop_coreloop_geometry}\textbf{(d)}), even as downstream performance can deteriorate.
This suggests that small per-iteration updates can still accumulate, gradually drifting recurrent states away from the distribution encountered during training.
Thus, increasingly small recurrent updates do not necessarily indicate that additional iterations remain useful; effective extrapolation also depends on how recurrent states evolve and how the shared core operates along this trajectory.
These observations motivate us to examine whether additional conditioning can help sustain useful recurrent computation as the trajectory evolves beyond the training horizon.
\vspace{-1.1mm}
\section{How Should Recurrence Be Conditioned?}
\label{sec:features}
\vspace{-1.2mm}

\begin{figure}[t]
    \centering
    \setlength{\abovecaptionskip}{5pt}
    \setlength{\belowcaptionskip}{-4pt}
    \vspace{-1.25mm}
    \includegraphics[width=1.0\linewidth]{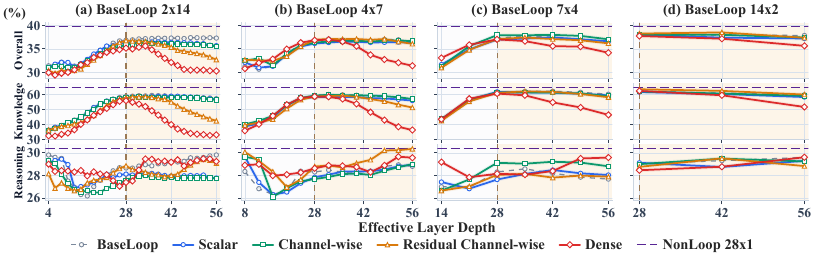}
    \caption{\footnotesize Initial-state injection results on Qwen3-0.6B BaseLoop configurations. BaseLoop provides the reference without state conditioning, and NonLoop $28\times1$ provides a non-recurrent reference.}
    \label{fig:initial_state_injection_qwen}
\end{figure}

We further investigate whether additional conditioning can mitigate performance degradation under mismatched training and inference budgets (Sec.~\ref{sec:comparison}). We consider \textit{State Conditioning}, which injects hidden-state information, and \textit{Timestep Conditioning}, which encodes the current recurrent iteration.

\vspace{-.6mm}
\subsection{State Conditioning}
\vspace{-.6mm}

\textbf{Initial-state injection and its limitations.}
We first study initial-state injection, which supplies the initial representation $h_0$ as a fixed reference at every recurrent iteration~\citep{huginn}:
\(h_{\ell+1}=F_\theta\!\left(\mathcal{I}_{\phi}(h_\ell,h_0)\right).\)
Huginn implements $\mathcal{I}_{\phi}$ by concatenating $h_\ell$ and $h_0$ followed by a linear projection (Dense).
To systematically examine how the form and capacity of this injection affect recurrent scaling, we additionally consider Scalar, Channel-wise, and Residual Channel-wise parameterizations:
\begin{equation}
\footnotesize
\begin{aligned}
\mathcal{I}_{\phi}^{\mathrm{scalar}}(h,h_0)
    &= h+\alpha h_0,
&
\;\mathcal{I}_{\phi}^{\mathrm{channel}}(h,h_0)
    &= a\odot h+b\odot h_0,\\
\mathcal{I}_{\phi}^{\mathrm{res}}(h,h_0)
    &= (\mathbf{1}+\delta_a)\odot h+b\odot h_0,
&
\;\mathcal{I}_{\phi}^{\mathrm{dense}}(h,h_0)
    &= W_hh+W_0h_0.
\end{aligned}
\label{eq:initial_state_injection_variants}
\end{equation}
Here, $\alpha\in\mathbb{R}$, $a,b,\delta_a\in\mathbb{R}^{d}$, and $W_h,W_0\in\mathbb{R}^{d\times d}$.
The maps act on the hidden dimension at every token position and are applied before each execution of the shared stack. 
The injection parameters are learned jointly with the backbone and initialized such that $\mathcal{I}_{\phi}(h,h_0)=h$.

As evaluated in Fig.~\ref{fig:initial_state_injection_qwen}, initial-state injection exhibits severe limitations, particularly on knowledge-intensive tasks. 
During loop extrapolation, the high-capacity Dense variant causes a catastrophic performance collapse in overall and knowledge accuracy across configurations, indicating that strong parametric conditioning overfits to the training loop count and disrupts knowledge retention during deep unrolling. 
In contrast, on reasoning tasks, performance remains largely flat across injection variants, showing minimal sensitivity to initial-state conditioning. 
Meanwhile, lightweight variants (Scalar, Channel-wise) avoid the severe breakdown on knowledge, but offer negligible net gains over the unconditioned BaseLoop baseline. 
Because Transformer architectures naturally retain input semantics via internal residual streams, forcibly re-injecting a static $h_0$ provides redundant semantics rather than dynamic trajectory guidance.

\begin{figure}[t]
    \centering
    \setlength{\abovecaptionskip}{5pt}
    \setlength{\belowcaptionskip}{-4pt}
    \vspace{-1.25mm}
    \includegraphics[width=1.0\linewidth]{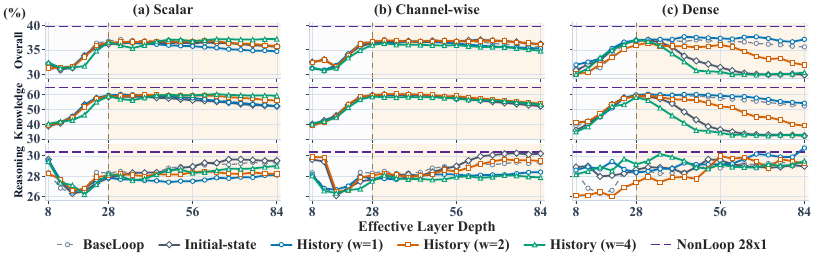}
    \caption{\footnotesize History-state injection on the Qwen3-0.6B BaseLoop $4$$\times$$7$ configuration. 
    BaseLoop gives the reference without injection, and
    NonLoop $28$$\times$$1$ is a non-recurrent reference.
    The dashed vertical line marks the effective training depth
    $L(K)=28$, and shading region indicates loop extrapolated up to $3\times$ the training depth.}
    \label{fig:history_state_injection_qwen}
\end{figure}

\textbf{History-state injection: conditioning on trajectory dynamics.}
While initial-state injection provides a static semantic anchor to where the recurrent trajectory originates, it fails to capture the evolving local dynamics during deep execution.
Driven by this limitation, we propose \textit{history-state injection}, which dynamically conditions on recent intermediate states.
To maintain numerical stability and smooth computation during inference extrapolation, we condition each update on \textit{relative state differences} rather than absolute historical vectors:
\begin{equation}
\footnotesize
h_{\ell+1}
=F_\theta\biggl(
h_\ell+\sum_{j=1}^{m_\ell}\mathcal{B}_j(h_{\ell-j}-h_\ell)
\biggr),
\qquad
m_\ell=\min\{w,\max(\ell-1,0)\}.
\label{eq:history_state_injection}
\end{equation}
This difference-based formulation explicitly models the local velocity of the recurrent path while preserving baseline recurrence when the history branch is initialized to zero. 
The history window includes only completed recurrent states $h_1,\ldots,h_{\ell-1}$, explicitly excluding $h_0$ to isolate history-state conditioning from static input injection. 
Each lag-specific operator $\mathcal{B}_j$ is shared across recurrent iterations and instantiated as a scalar, channel-wise vector, or dense linear map.

We evaluate history-state injection across different window sizes ($w \in \{1, 2, 4\}$) on the $4\times7$ BaseLoop configuration in Fig.~\ref{fig:history_state_injection_qwen}. 
We find that history conditioning acts as a powerful dynamic regularizer for high-capacity mappings. Under the dense parameterization (Fig.~\ref{fig:history_state_injection_qwen}\textbf{(c)}), history conditioning with a minimal window ($w=1$) rescues the model from static injection collapse, sustaining robust accuracy near the NonLoop reference. 
Two observations underpin this stabilizing effect:
\begin{itemize}[leftmargin=*,itemsep=0pt,topsep=2pt]
    \item \textit{Window-size sensitivity under Dense mapping}: Expanding the memory window to $w=2$ or $w=4$ progressively diminishes extrapolation performance, indicating that a minimal one-step difference ($\Delta h_{k-1}$) provides sufficient velocity cues without introducing redundant temporal noise.
    \item \textit{Negligible gains under low-capacity mappings}: For Scalar and Channel-wise variants (Fig.~\ref{fig:history_state_injection_qwen}\textbf{(a,b)}), history injection provides limited gains over BaseLoop and slightly hurts reasoning performance.
\end{itemize}

\vspace{-.5mm}
\subsection{Timestep Conditioning}
\vspace{-.5mm}

Inspired by~\citet{expressive_power,loopformer}, we condition the recurrent computation on normalized timestep size using three variants.
Loop Gating (LG) uses a time-dependent scalar to scale
the difference between the shared stack's output and its input.
Branch Gating (BG) applies separate time-dependent
scalar gates to the attention and MLP residual branches
within each layer.
\textit{AdaLN} adopts LoopFormer's modulation
structure~\citep{loopformer}, applying channel-wise scaling
to the RMSNorm outputs and channel-wise gating to the
corresponding residual branches.
All three variants use the same continuous time features.
At inference, we rescale the time grid to the requested number of
recurrent iterations: for $L_{\mathrm{infer}}$ iterations, iteration $\ell$ receives normalized time $t_\ell=\ell/L_{\mathrm{infer}}$
and step size $\Delta t=1/L_{\mathrm{infer}}$
($\ell=0,\ldots,L_{\mathrm{infer}}-1$).
Increasing the inference budget therefore divides the same
normalized time interval $[0,1]$ into more, smaller steps.
Full results are provided in App.~\ref{app:timestep_conditioning_results}.

\begin{figure}[t!]
    \centering
    \setlength{\abovecaptionskip}{5pt}
    \setlength{\belowcaptionskip}{-4pt}
    \vspace{-1mm}
    \includegraphics[width=1.0\linewidth]{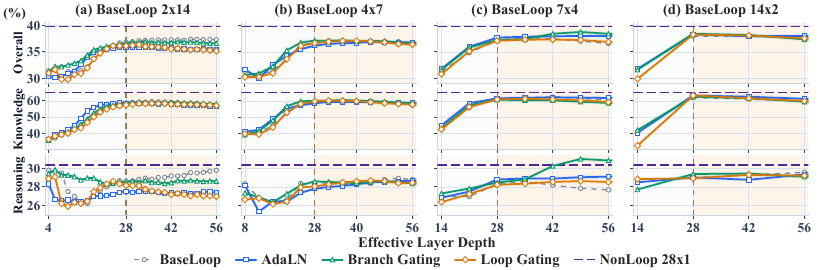}
        \caption{\footnotesize Timestep conditioning results of the Qwen3-0.6B backbone. All timestep variants are compared using rescaled time grids at inference. NonLoop $28\times1$ provides an unshared reference.} 
    \label{fig:timestep_conditioning_qwen}
\end{figure}

\textbf{Timestep conditioning enhances recurrent scaling, with efficacy tied to architecture design.}
As shown in Fig.~\ref{fig:timestep_conditioning_qwen}, LG and BG variants can improve overall performance at extrapolation parts, most notably in the $7\times4$ setup.
Specifically at $L(r)=56$, BG and LG achieve overall scores of 38.46 and 36.91, respectively, outperforming BaseLoop (36.59, Fig.~\ref{fig:timestep_conditioning_qwen}\textbf{c}).
However, the same schemes provide much smaller or negative gains
at this budget for $4\times7$ and $2\times14$
(Fig.~\ref{fig:timestep_conditioning_qwen}\textbf{(a,b)}).
The $14\times2$ results likewise show that improvements at the
training depth do not guarantee a consistent advantage at larger
budgets (Fig.~\ref{fig:timestep_conditioning_qwen}\textbf{(d)}).
Timestep-dependent gating therefore interacts with the allocation
of shared-stack depth and recurrent iterations.
Its effectiveness must be assessed jointly with the recurrent
architecture and the intended inference budget.

\textbf{Loop and Branch gating provide lightweight alternatives to AdaLN.}
Our proposed LG and BG use scalar modulation at the loop or
residual-branch level, requiring fewer conditioning parameters than channel-wise AdaLN modulation.
For $7\times4$ at $L(r)=56$, BG outperforms AdaLN in both
overall (38.46 vs 38.01) and reasoning (30.88 vs 29.12)
(Fig.~\ref{fig:timestep_conditioning_qwen}\textbf{(c)}).
LG also achieves a higher Reasoning score for $14\times2$
at $L(r)=56$ (29.61 vs 28.78)
(Fig.~\ref{fig:timestep_conditioning_qwen}\textbf{(d)}).
These results demonstrate that simple scalar modulation can compete with more expressive channel-wise conditioning, and that the preferred modulation granularity depends on the recurrent architecture.
AdaLN~\citep{loopformer} better preserves Knowledge in these comparisons, indicating a task-dependent trade-off.
LG and BG therefore provide lightweight design options for controlling recurrent computation.
These results motivate combining timestep conditioning with input injection to test whether knowledge retention and reasoning gains can be maintained jointly.
\vspace{-1mm}
\section{When Are Conditioning Mechanisms Composable?}
\label{sec:mechanism}
\vspace{-1mm}

The preceding results show that state and timestep conditioning improve recurrent scaling through distinct signals, motivating us to examine whether their benefits are complementary.
Using Qwen3-0.6B BaseLoop $4\times7$, we evaluate all pairwise combinations of initial-state, history-state, and timestep conditioning, as well as the combination of all three, up to an effective depth of $84$ (3x training depth).

\begin{figure}[t!]
    \centering
    \setlength{\abovecaptionskip}{4pt}
    \setlength{\belowcaptionskip}{2pt}
    \vspace{-1.25mm}
    \includegraphics[width=1.0\linewidth]{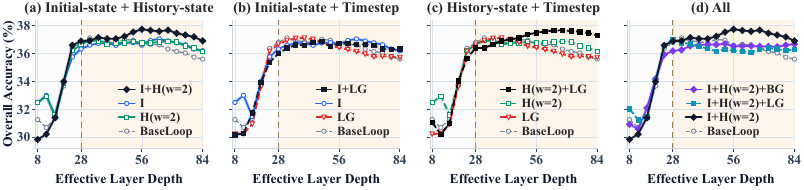}
    \caption{\footnotesize Performance of combined conditioning mechanisms for Qwen3-0.6B BaseLoop $4\times7$. All state conditioning uses the channel-wise parameterization, with history window  \(w=2\). In the legends, I and H denote initial-state and history-state injection, LG denotes loop gating, and BG denotes branch gating.}
    \label{fig:combine_results}
\end{figure}

\begin{figure}[t]
    \centering
    \setlength{\abovecaptionskip}{5pt}
    \setlength{\belowcaptionskip}{-4pt}
    \includegraphics[width=1.0\linewidth]{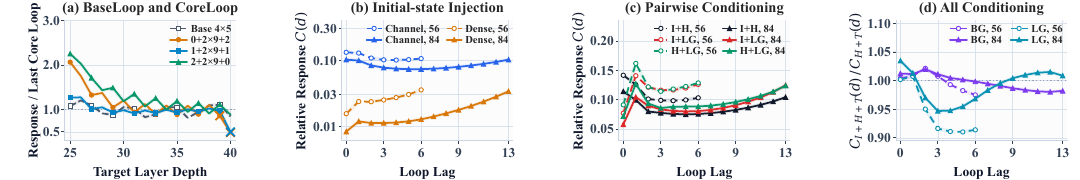}
   \caption{\footnotesize Results of relative update responses to skipping an earlier block occurrence.
    \textbf{(a)} Llama BaseLoop/CoreLoop at $D=40$: responses averaged over separate skips at positions 21-24, divided by each model's mean response in its final recurrent loop. Crosses mark coda blocks.
    \textbf{(b,c)} Qwen BaseLoop $4\times7$: mean $C$ grouped by loop lag for initial-state injection and pairwise conditioning.
    \textbf{(d)} Three-mechanism responses divided by the corresponding H+T responses, with the timestep and history settings matched.
    All state injections in \textbf{(c,d)} are channel-wise, with $w=2$ for history.
    In \textbf{(b-d)}, dashed curves with open circles denote $D=56$ and solid
    curves with filled triangles denote $D=84$; all measured lags are shown.}
    \label{fig:conditioning_responses}
\end{figure}

\textbf{History-state and timestep conditioning form the strongest pair under deep extrapolation.}
At $D=84$, H+LG reaches 37.30 Overall accuracy, exceeding I+H (36.91) and I+LG (36.33), as well as its stronger individual component by 1.16 percentage points (Fig.~\ref{fig:combine_results}(\textbf{a,b,c})). 
Adding initial-state conditioning provides no further benefit, with the three-way combinations remaining below H+LG (Fig.~\ref{fig:combine_results}\textbf{(d)}).
These results identify history-state and timestep conditioning as the strongest complementary pair for recurrent scaling, with no additional gains from initial-state conditioning.

\textbf{Geometry analysis.} 
To understand this complementarity beyond performance, we examine how earlier computation influences later recurrent updates. 
We measure \textit{computational interaction} $C$, defined as the relative change in a downstream block's update after skipping an earlier block occurrence (App.~\ref{app:probing}).
Unlike angular distance or update magnitude, which quantify how much a representation changes, $C$ captures how strongly later computation depends on earlier computation.
\begin{itemize}[leftmargin=*,itemsep=0pt,topsep=2pt]
    \item \textit{Computational interaction reveals information beyond update magnitude.}
    Large state transformations need not imply strong computational interaction (Fig.~\ref{fig:conditioning_responses}\textbf{(a,b)}): Coda layers reduce the terminal-to-core response ratio from $0.85$-$0.89$ to $0.48$-$0.50$, and channel-wise initial-state injection shows nearly $7\times$ the cross-loop response of Dense injection at $D=84$.
    Thus, transformation magnitude alone does not characterize effective iterative refinement.
    \item \textit{History-state and timestep complementarity emerges through cross-loop interaction.}
    At $D=84$, H+LG exhibits $18.6\%$ and $10.4\%$ higher mean interaction over lags $1$--$6$ than I+H and I+LG, respectively, consistent with its higher accuracy (Fig.~\ref{fig:combine_results}).
    This advantage does not come from uniformly larger responses: H+LG shows a smaller within-loop response than I+H, with its gain appearing mainly across loops (Fig.~\ref{fig:conditioning_responses}\textbf{(c)}).
    This suggests complementary roles across iterations: history-state conditioning exposes how previous states evolve, while timestep conditioning modulates the shared update at each iteration.
    Together, they allow later iterations to build more strongly on earlier computation, whereas pairings with the static initial state provide weaker cross-loop coupling.
    \item \textit{Initial-state conditioning provides no additional cross-loop benefit.}
    Adding initial-state conditioning to H+LG preserves local interactions but reduces the mean interaction over lags $2$-$6$ by $7.4\%$ and $4.2\%$ at $D=56$ and $84$, respectively (Fig.~\ref{fig:conditioning_responses}\textbf{(d)}), alongside lower accuracy for the three-way combination.
    The same trend appears in the unnormalized responses, indicating that it is not solely a normalization effect.
    Although the pattern is not universal across all lags and gating variants, it provides a possible explanation for the lack of additive gains from initial-state conditioning once history-state and timestep conditioning are combined.
\end{itemize}
Overall, these results suggest that effective recurrent scaling depends not only on the magnitude of state transformations, but also on how computation interacts across recurrent iterations.

\vspace{-1mm}
\section*{Acknowledgement}
\vspace{-1mm}
The work described in this paper was supported by the General Research Fund and Early Career Scheme by the Research Grants Council of Hong Kong (Project Number: 24211626).

\bibliography{iclr2027_conference}

@article{llama3,
  title={The llama 3 herd of models},
  author={Grattafiori, Aaron and Dubey, Abhimanyu and Jauhri, Abhinav and Pandey, Abhinav and Kadian, Abhishek and Al-Dahle, Ahmad and Letman, Aiesha and Mathur, Akhil and Schelten, Alan and Vaughan, Alex and others},
  journal={arXiv preprint arXiv:2407.21783},
  url={https://arxiv.org/abs/2407.21783},
  year={2024}
}

@article{qwen3,
  title={Qwen3 technical report},
  author={Yang, An and Li, Anfeng and Yang, Baosong and Zhang, Beichen and Hui, Binyuan and Zheng, Bo and Yu, Bowen and Gao, Chang and Huang, Chengen and Lv, Chenxu and others},
  journal={arXiv preprint arXiv:2505.09388},
  url={https://arxiv.org/abs/2505.09388},
  year={2025}
}

@misc{muon,
  author       = {Keller Jordan and Yuchen Jin and Vlado Boza and You Jiacheng and
                  Franz Cesista and Laker Newhouse and Jeremy Bernstein},
  title        = {Muon: An optimizer for hidden layers in neural networks},
  year         = {2024},
  url          = {https://kellerjordan.github.io/posts/muon/}
}

@article{dmuon,
  title={Muon is scalable for llm training},
  author={Liu, Jingyuan and Su, Jianlin and Yao, Xingcheng and Jiang, Zhejun and Lai, Guokun and Du, Yulun and Qin, Yidao and Xu, Weixin and Lu, Enzhe and Yan, Junjie and others},
  journal={arXiv preprint arXiv:2502.16982},
  url={https://arxiv.org/abs/2502.16982},
  year={2025}
}

@inproceedings{adamw,
    title={Decoupled Weight Decay Regularization},
    author={Loshchilov, Ilya and Hutter, Frank},
    booktitle={ICLR},
    year={2019},
    url={https://openreview.net/forum?id=Bkg6RiCqY7}
}

@inproceedings{baseloop,
  title={Reasoning with latent thoughts: On the power of looped transformers},
  author={Saunshi, Nikunj and Dikkala, Nishanth and Li, Zhiyuan and Kumar, Sanjiv and J Reddi, Sashank},
  booktitle={ICLR},
  year={2025},
  url={https://openreview.net/forum?id=din0lGfZFd}
}

@article{ouro,
  title={Scaling latent reasoning via looped language models},
  author={Zhu, Rui-Jie and Wang, Zixuan and Hua, Kai and Zhang, Tianyu and Li, Ziniu and Que, Haoran and Wei, Boyi and Wen, Zixin and Yin, Fan and Xing, He and others},
  journal={arXiv preprint arXiv:2510.25741},
  url={https://arxiv.org/abs/2510.25741}, 
  year={2025}
}

@inproceedings{mor,
    title={Mixture-of-Recursions: Learning Dynamic Recursive Depths for Adaptive Token-Level Computation},
    author={Sangmin Bae and Yujin Kim and Reza Bayat and Sungnyun Kim and Jiyoun Ha and Tal Schuster and Adam Fisch and Hrayr Harutyunyan and Ziwei Ji and Aaron Courville and Se-Young Yun},
    booktitle={NeurIPS},
    year={2025},
    url={https://openreview.net/forum?id=QuqsEIVWIG}
}

@inproceedings{huginn,
  title={Scaling up test-time compute with latent reasoning: A recurrent depth approach},
  author={Geiping, Jonas and McLeish, Sean and Jain, Neel and Kirchenbauer, John and Singh, Siddharth and Bartoldson, Brian and Kailkhura, Bhavya and Bhatele, Abhinav and Goldstein, Tom},
  booktitle={NeurIPS},
  year={2025},
  url={https://openreview.net/forum?id=S3GhJooWIC}
}

@inproceedings{loopformer,
    title={LoopFormer: Elastic-Depth Looped Transformers for Latent Reasoning via Shortcut Modulation},
    author={Ahmadreza Jeddi and Marco Ciccone and Babak Taati},
    booktitle={ICLR},
    year={2026},
    url={https://openreview.net/forum?id=RzYXb5YWBs}
}

@article{fprm,
  title={Fixed-Point Reasoners: Stable and Adaptive Deep Looped Transformers},
  author={Movahedi, Sajad and Milovanovi{\'c}, Vera and Feigin, Shlomo Libo and Theus, Alexander and Hofmann, Thomas and Boeva, Valentina and Rusch, T Konstantin and Orvieto, Antonio},
  journal={arXiv preprint arXiv:2606.18206},
  url={https://arxiv.org/abs/2606.18206}, 
  year={2026}
}

@article{deeploop,
  title={DeepLoop: Depth Scaling for Looped Transformers},
  author={Li, Shuzhen and Zhang, Yifan and Guo, Jiacheng and Gu, Quanquan and Wang, Mengdi},
  journal={arXiv preprint arXiv:2607.13491},
  url={https://arxiv.org/abs/2607.13491}, 
  year={2026}
}

@article{stars,
  title={Stabilizing recurrent dynamics for test-time scalable latent reasoning in looped language models},
  author={Yang, Xiao-Wen and Han, Ziyu and Zhang, Xi-Hua and Wei, Wen-Da and Shao, Jie-Jing and Guo, Lan-Zhe and Li, Yu-Feng},
  journal={arXiv preprint arXiv:2605.26733},
  url={https://arxiv.org/abs/2605.26733}, 
  year={2026}
}

@article{yocou,
  title={Universal YOCO for Efficient Depth Scaling},
  author={Sun, Yutao and Dong, Li and Ye, Tianzhu and Huang, Shaohan and Wang, Jianyong and Wei, Furu},
  journal={arXiv preprint arXiv:2604.01220},
  url={https://arxiv.org/abs/2604.01220}, 
  year={2026}
}

@article{LT2,
  title={LT2: Linear-Time Looped Transformers},
  author={Deng, Chunyuan and Zhang, Yizhe and Zhu, Rui-Jie and Xu, Yuanyuan and Liu, Jiarui and Ng, TS and Chen, Hanjie},
  journal={arXiv preprint arXiv:2605.20670},
  url={https://arxiv.org/abs/2605.20670}, 
  year={2026}
}

@inproceedings{rins,
    title={Recursive Inference Scaling: A Winning Path to Scalable Inference in Language and Multimodal Systems},
    author={Ibrahim Alabdulmohsin and Xiaohua Zhai},
    booktitle={NeurIPS},
    year={2025},
    url={https://openreview.net/forum?id=cLbGkINOLP}
}

@article{attractor,
  title={Solve the Loop: Attractor Models for Language and Reasoning},
  author={Fein-Ashley, Jacob and Rashidinejad, Paria},
  journal={arXiv preprint arXiv:2605.12466},
  url={https://arxiv.org/abs/2605.12466}, 
  year={2026}
}

@inproceedings{polar,
    title={Skip a Layer or Loop It? Learning Program-of-Layers in {LLM}s},
    author={Ziyue Li and Yang Li and Tianyi Zhou},
    booktitle={ICML},
    year={2026},
    url={https://openreview.net/forum?id=pl10b6EQAN}
}

@article{loopmoe,
  title={Sparse layers are critical to scaling looped language models},
  author={Lee, Ryan and Biloki, Jacob and Hu, Edward J and May, Jonathan},
  journal={arXiv preprint arXiv:2605.09165},
  url={https://arxiv.org/abs/2605.09165}, 
  year={2026}
}

@article{mixerloop,
  title={Allocating Recurrent Compute in Looped Language Models},
  author={Lin, Ruhai and Guo, Yiyang and Zhu, Rui-Jie and Ye, Hao and Eshraghian, Jason K},
  journal={arXiv preprint arXiv:2608.18230},
  url={https://arxiv.org/abs/2608.18230}, 
  year={2026}
}

@article{retrofitting,
  title={Teaching pretrained language models to think deeper with retrofitted recurrence},
  author={McLeish, Sean and Li, Ang and Kirchenbauer, John and Kalra, Dayal Singh and Bartoldson, Brian R and Kailkhura, Bhavya and Schwarzschild, Avi and Geiping, Jonas and Goldstein, Tom and Goldblum, Micah},
  journal={arXiv preprint arXiv:2511.07384},
  url={https://arxiv.org/abs/2511.07384}, 
  year={2025}
}

@article{ETD,
  title={Encode, Think, Decode: Scaling test-time reasoning with recursive latent thoughts},
  author={Koishekenov, Yeskendir and Lipani, Aldo and Cancedda, Nicola},
  journal={arXiv preprint arXiv:2510.07358},
  url={https://arxiv.org/abs/2510.07358}, 
  year={2025}
}

@article{ouroboros,
  title={Ouroboros: Dynamic Weight Generation for Recursive Transformers via Input-Conditioned LoRA Modulation},
  author={Jaber, Jaber and Jaber, Osama},
  journal={arXiv preprint arXiv:2604.02051},
  url={https://arxiv.org/abs/2604.02051}, 
  year={2026}
}

@article{retrofitting_recurrence,
  title={Retrofitting Recurrent Depth into a Pretrained Language Model: Installation, Extrapolation, Transfer, and Retention at Two Parameter Budgets},
  author={Shapiro, Mark},
  journal={arXiv preprint arXiv:2608.11233},
  url={https://arxiv.org/abs/2608.11233}, 
  year={2026}
}

@article{mechanistic,
  title={A mechanistic analysis of looped reasoning language models},
  author={Blayney, Hugh and Arroyo, {\'A}lvaro and Obando-Ceron, Johan and Castro, Pablo Samuel and Courville, Aaron and Bronstein, Michael M and Dong, Xiaowen},
  journal={arXiv preprint arXiv:2604.11791},
  url={https://arxiv.org/abs/2604.11791}, 
  year={2026}
}

@inproceedings{LTG,
    title={Loop, Think, \& Generalize: Implicit Reasoning in Recurrent-Depth Transformers},
    author={Harsh Kohli and Srinivasan Parthasarathy and Huan Sun and Yuekun Yao},
    booktitle={CoLM},
    year={2026},
    url={https://openreview.net/forum?id=8fz7WRThKL}
}

@inproceedings{expressive_power,
    title={On Expressive Power of Looped Transformers: Theoretical Analysis and Enhancement via Timestep Encoding},
    author={Kevin Xu and Issei Sato},
    booktitle={ICML},
    year={2025},
    url={https://openreview.net/forum?id=H4BuhRezCV}
}

@article{adaptive_depth,
  title={Adaptive Depth in Looped Transformers: Diagnosing Learned Halting Gates and Trajectory Readouts},
  author={Popescu, Andrei Cristian and Borde, Haitz S{\'a}ez de Oc{\'a}riz and Li{\`o}, Pietro},
  journal={arXiv preprint arXiv:2607.20519},
  url={https://arxiv.org/abs/2607.20519}, 
  year={2026}
}

@article{token_FP,
  title={Per-Token Fixed-Point Convergence in Depth-Recurrent Transformers},
  author={Logan, Joe},
  journal={arXiv preprint arXiv:2607.14427},
  url={https://arxiv.org/abs/2607.14427}, 
  year={2026}
}

@article{residual_scaling,
  title={On the Residual Scaling of Looped Transformers: Stability and Transferability},
  author={Wang, Shaowen and Li, Bingrui and Zhang, Ge and Huang, Wenhao and Yan, Shen and Li, Jian},
  journal={arXiv preprint arXiv:2606.18524},
  url={https://arxiv.org/abs/2606.18524}, 
  year={2026}
}

@article{parcae,
  title={Parcae: Scaling laws for stable looped language models},
  author={Prairie, Hayden and Novack, Zachary and Berg-Kirkpatrick, Taylor and Fu, Daniel Y},
  journal={arXiv preprint arXiv:2604.12946},
  url={https://arxiv.org/abs/2604.12946}, 
  year={2026}
}

@article{how_much_scaling,
  title={How much is one recurrence worth? iso-depth scaling laws for looped language models},
  author={Schwethelm, Kristian and Rueckert, Daniel and Kaissis, Georgios},
  journal={arXiv preprint arXiv:2604.21106},
  url={https://arxiv.org/abs/2604.21106}, 
  year={2026}
}

@article{smelt,
  title={SMELT: Scaling Laws for Compute-Matched MoE Looped Transformers},
  author={Wang, Shaowen and Zhang, Ge and Luo, Kairong and Wu, Yuhao and Liu, Shaofan and Liu, Jiaheng and Huang, Wenhao and Yan, Shen and Li, Jian},
  journal={arXiv preprint arXiv:2609.01343},
  url={https://arxiv.org/abs/2609.01343}, 
  year={2026}
}

@inproceedings{fineweb,
     author = {Penedo, Guilherme and Kydl\'{\i}\v{c}ek, Hynek and allal, Loubna Ben and Lozhkov, Anton and Mitchell, Margaret and Raffel, Colin and Von Werra, Leandro and Wolf, Thomas},
     booktitle = {NeurIPS},
     title = {The FineWeb Datasets: Decanting the Web for the Finest Text Data at Scale},
     url = {https://proceedings.neurips.cc/paper_files/paper/2024/file/370df50ccfdf8bde18f8f9c2d9151bda-Paper-Datasets_and_Benchmarks_Track.pdf},
     year = {2024}
}

@article{arc,
  title={Think you have solved question answering? try arc, the ai2 reasoning challenge},
  author={Clark, Peter and Cowhey, Isaac and Etzioni, Oren and Khot, Tushar and Sabharwal, Ashish and Schoenick, Carissa and Tafjord, Oyvind},
  journal={arXiv preprint arXiv:1803.05457},
  year={2018}
}

@inproceedings{sciq,
    title = "Crowdsourcing Multiple Choice Science Questions",
    author = "Welbl, Johannes  and
      Liu, Nelson F.  and
      Gardner, Matt",
    booktitle = "Proceedings of the 3rd Workshop on Noisy User-generated Text",
    year = "2017",
    url = "https://aclanthology.org/W17-4413"
 }

@inproceedings{hellaswag,
    title = "{H}ella{S}wag: Can a Machine Really Finish Your Sentence?",
    author = "Zellers, Rowan  and
      Holtzman, Ari  and
      Bisk, Yonatan  and
      Farhadi, Ali  and
      Choi, Yejin",
    booktitle = "ACL",
    year = "2019",
    url = "https://aclanthology.org/P19-1472"
 }

@inproceedings{winogrande,
  title={WinoGrande: An Adversarial Winograd Schema Challenge at Scale},
  author={Sakaguchi, Keisuke and Le Bras, Ronan and Bhagavatula, Chandra and Choi, Yejin},
  booktitle={AAAI},
  year={2020}
}

@inproceedings{commonsenseqa,
    title = "{C}ommonsense{QA}: A Question Answering Challenge Targeting Commonsense Knowledge",
    author = "Talmor, Alon  and
      Herzig, Jonathan  and
      Lourie, Nicholas  and
      Berant, Jonathan",
    booktitle = "NAACL-HLT",
    year = "2019",
    url = "https://aclanthology.org/N19-1421"
}

@inproceedings{openbookqa,
    title = "Can a Suit of Armor Conduct Electricity? A New Dataset for Open Book Question Answering",
    author = "Mihaylov, Todor  and
      Clark, Peter  and
      Khot, Tushar  and
      Sabharwal, Ashish",
    booktitle = "EMNLP",
    year = "2018",
    url = "https://aclanthology.org/D18-1260"
}

@inproceedings{piqa,
  author    = {Yonatan Bisk and Rowan Zellers and Ronan Le Bras and Jianfeng Gao and Yejin Choi},
  title     = {PIQA: Reasoning about Physical Commonsense in Natural Language},
  booktitle = {AAAI},
  year      = {2020},
  url       = {https://ojs.aaai.org/index.php/AAAI/article/view/6239}
}

@inproceedings{proofwriter,
    title = "{P}roof{W}riter: Generating Implications, Proofs, and Abductive Statements over Natural Language",
    author = "Tafjord, Oyvind  and
      Dalvi, Bhavana  and
      Clark, Peter",
    booktitle = "Findings of ACL",
    year = "2021",
    url = "https://aclanthology.org/2021.findings-acl.317/"
}

@inproceedings{clutrr,
    title = "{CLUTRR}: A Diagnostic Benchmark for Inductive Reasoning from Text",
    author = "Sinha, Koustuv  and
      Sodhani, Shagun  and
      Dong, Jin  and
      Pineau, Joelle  and
      Hamilton, William L.",
    booktitle = "Proceedings of the 2019 Conference on Empirical Methods in Natural Language Processing and the 9th International Joint Conference on Natural Language Processing (EMNLP-IJCNLP)",
    month = nov,
    year = "2019",
    address = "Hong Kong, China",
    publisher = "Association for Computational Linguistics",
    url = "https://aclanthology.org/D19-1458/",
    doi = "10.18653/v1/D19-1458",
    pages = "4506--4515",
}

@inproceedings{bbh,
    title = "Challenging {BIG}-Bench Tasks and Whether Chain-of-Thought Can Solve Them",
    author = {Suzgun, Mirac  and
      Scales, Nathan  and
      Sch{\"a}rli, Nathanael  and
      Gehrmann, Sebastian  and
      Tay, Yi  and
      Chung, Hyung Won  and
      Chowdhery, Aakanksha  and
      Le, Quoc  and
      Chi, Ed  and
      Zhou, Denny  and
      Wei, Jason},
    booktitle = "Findings of ACL",
    year = "2023",
    url = "https://aclanthology.org/2023.findings-acl.824/"
}

@inproceedings{mmlu,
    title={Measuring Massive Multitask Language Understanding},
    author={Dan Hendrycks and Collin Burns and Steven Basart and Andy Zou and Mantas Mazeika and Dawn Song and Jacob Steinhardt},
    booktitle={ICLR},
    year={2021},
    url={https://openreview.net/forum?id=d7KBjmI3GmQ}
}

@inproceedings{race,
    title = "{RACE}: Large-scale {R}e{A}ding Comprehension Dataset From Examinations",
    author = "Lai, Guokun  and
      Xie, Qizhe  and
      Liu, Hanxiao  and
      Yang, Yiming  and
      Hovy, Eduard",
    booktitle = "EMNLP",
    year = "2017",
    url = "https://aclanthology.org/D17-1082"
}

@misc{eval-harness,
  author       = {Gao, Leo and Tow, Jonathan and Abbasi, Baber and Biderman, Stella and Black, Sid and DiPofi, Anthony and Foster, Charles and Golding, Laurence and Hsu, Jeffrey and Le Noac'h, Alain and Li, Haonan and McDonell, Kyle and Muennighoff, Niklas and Ociepa, Chris and Phang, Jason and Reynolds, Laria and Schoelkopf, Hailey and Skowron, Aviya and Sutawika, Lintang and Tang, Eric and Thite, Anish and Wang, Ben and Wang, Kevin and Zou, Andy},
  title        = {The Language Model Evaluation Harness},
  month        = 07,
  year         = 2024,
  publisher    = {Zenodo},
  version      = {v0.4.3},
  url          = {https://zenodo.org/records/12608602}
}
\bibliographystyle{iclr2027_conference}

\newpage
\appendix
\onecolumn
\addcontentsline{toc}{section}{Appendix}
\renewcommand \thepart{}
\renewcommand \partname{}
\part{\Large{\centerline{Appendix}}}
\parttoc

\newpage

\section{Limitations}

Our study has several limitations.
First, due to computational constraints, our experiments are conducted on models of up to roughly 1B parameters (Llama3.1-1B and Qwen3-0.6B configurations) under a fixed token budget. 
Second, we evaluate extrapolation up to a fixed multiple of the training horizon with a manually specified loop count, and do not explore adaptive mechanisms that determine the number of iterations per input.
Finally, our geometry analyses, including computational interaction, are intended to offer complementary insights into recurrent dynamics.
Further study is needed to more fully establish their relationship with downstream performance.

\section{Extended Literature Review}

\paragraph{LoopLM variants.}

Looped language models (LoopLMs) increase effective depth by repeatedly applying a shared set of Transformer layers, thereby decoupling sequential computation from the number of unique parameters. 
Existing LoopLM variants can be broadly divided into two categories according to how recurrence is introduced: models that are pre-trained with recurrent depth \textit{from scratch}, and methods that \textit{retrofit} recurrence into pretrained Transformers through continual pre-training.

\textbf{Strategy 1: Pre-training from scratch.}

Early works such as BaseLoop~\cite{baseloop} and Huginn~\cite{huginn} establish recurrent depth as a viable alternative to conventional depth scaling, showing that repeatedly applying shared layers can provide additional latent computation and improve reasoning tasks without proportionally increasing parameter count. 
Subsequent methods focus on making recurrence more adaptive and scalable. 
Ouro~\cite{ouro}, MoR~\cite{mor}, and LoopFormer~\cite{loopformer} introduce dynamic or elastic recurrence, allowing computation depth to vary across inputs, tokens, or inference budgets. 
Other works address the stability of deep recurrence: FPRM~\cite{fprm}, DeepLoop~\cite{deeploop}, and STARS~\cite{stars} study fixed-point behaviors, residual scaling, and recurrent dynamical stability, respectively, with the goal of enabling reliable extrapolation to larger loop counts than those seen during training.

A parallel line explores which computations should be repeated and how recurrent architectures can be made more efficient or expressive. 
RINS~\cite{rins} studies alternative recursive execution patterns and identifies $A^rB$ as an effective topology, while YOCO-U~\cite{yocou} and LT2~\cite{LT2} combine recurrent depth with efficient attention or KV-cache designs. 
Attractor models~\cite{attractor} formulate iterative latent refinement as convergence toward an equilibrium, providing an implicit alternative to finite-depth recurrence. 
More recent works further examine the granularity of looping.
Looped-MoE~\cite{loopmoe} shows that sparse expert routing can increase functional diversity across repeated passes, while MixerLoop~\cite{mixerloop} demonstrates that recurrent compute can be concentrated on selected Transformer submodules.
PoLar~\cite{polar} similarly broadens the design space by learning flexible programs that can skip or repeat different layers instead of using a fixed recurrent block.

\textbf{Strategy 2: Retrofitting through continual pre-training.}

Instead of introducing recurrence during initial pre-training, another line of work converts existing pretrained Transformers into recurrent models. 
ETD~\citep{ETD} identifies a small subset of reasoning-relevant intermediate layers and trains the model to repeatedly apply these layers as a latent stage, enabling additional inference-time computation while preserving the original parameter count and overall architecture. 
~\citet{retrofitting} more systematically studies the conversion of standard pretrained Transformers into depth-recurrent models, showing that a curriculum that gradually increases the number of recurrences can preserve pretrained capabilities while efficiently adapting the model to repeated weight reuse. Ouroboros~\cite{ouroboros} addresses a limitation of such weight sharing, namely that identical recurrent weights repeatedly apply the same transformation, by using an input-conditioned controller to dynamically modulate LoRA parameters at each recurrence step, allowing shared layers to implement step- and input-dependent transformations. 
More recently, \citet{retrofitting_recurrence} explores the installation and persistence of recurrent computation in pretrained models in greater detail, showing that continual training can induce reusable iterative latent procedures that extrapolate beyond the supervised recurrence depth and persist under subsequent outcome-only training. 

\paragraph{Understanding LoopLMs.}

A growing body of work analyzes the internal dynamics and computational properties of LoopLMs. 
\citet{mechanistic} shows that repeated computation gives rise to structured recurrent trajectories, where hidden states and attention patterns progressively stabilize across loops and exhibit fixed-point-like behaviors. 
Building on this perspective, \citet{token_FP} finds that convergence is strongly token-dependent, with different tokens reaching stable representations after different numbers of recurrent steps. 
\citet{adaptive_depth} further examines how such convergence can be exploited for adaptive computation, showing that the effectiveness of learned halting depends not only on the stopping criterion itself but also on how training shapes the recurrent trajectory and final readout.

Other studies investigate the theoretical and optimization properties induced by recurrence. 
\citet{expressive_power} characterizes the expressive limitations of weight-shared Transformers and show that timestep encoding can increase the functional diversity across recurrent steps. 
From a generalization perspective, \citet{LTG} demonstrates on synthetic controlled compositional tasks that recurrence promotes systematic generalization and depth extrapolation, although excessive looping can lead to overthinking. 
Finally, \citet{residual_scaling} analyzes signal propagation under repeated weight reuse and derive loop-specific residual scaling rules, highlighting that stability conditions for recurrent depth differ fundamentally from those of conventional Transformers with independent layers. 

\paragraph{Scaling laws of LoopLMs.}

Recent work has begun to characterize recurrent depth as a distinct scaling dimension in LoopLMs. 
Parcae~\citep{parcae} derives scaling laws over model size, training data, and recurrence depth, showing that recurrence should scale jointly with data under fixed compute and that test-time depth exhibits diminishing returns. 
Complementarily, \citet{how_much_scaling} quantify the value of recurrence relative to unique depth, estimating a recurrence-equivalence exponent of $\varphi=0.46$, which indicates that repeated layers provide additional capacity but are substantially less effective than independent layers. 
More recently, SMELT~\citep{smelt} extends this analysis to MoE-based LoopLMs under a stricter budget-matched setting, jointly controlling per-token FLOPs, non-embedding parameters, and KV cache. 
By fitting separate Chinchilla-style scaling laws up to 54B non-embedding parameters, SMELT shows that looped MoE models scale more favorably than their unlooped counterparts, requiring 6.8--18.0\% fewer training FLOPs along the compute-optimal frontier.

\newpage
\section{Experimental Settings}
\label{app:exp_settings}

\subsection{Models and Data}
\label{app:models_and_data}

\paragraph{Backbone architectures.}

We construct all models from scratch using two decoder-only Transformer architectures: a 20-layer Llama 3.1-style model with approximately one billion parameters, denoted Llama3.1-1B, and the 28-layer Qwen3-0.6B architecture.
These names refer to the corresponding non-recurrent reference configurations. 
The recurrent models introduced below preserve their block-level architectural dimensions while instantiating fewer physical Transformer blocks.
Both architectures use pre-normalization (Pre-LN), causal self-attention, grouped-query
attention (GQA), rotary positional embeddings (RoPE), and SwiGLU feed-forward
networks. 
Each Transformer block contains an RMSNorm before the attention sublayer and another RMSNorm before the feed-forward sublayer. 
Attention and MLP projections are bias-free, and we use zero attention dropout.

The Llama3.1-1B reference configuration contains 20 Transformer blocks with model dimension \(d_{\mathrm{model}}=1536\) and feed-forward dimension \(d_{\mathrm{ff}}=5376\). 
Its attention module has 12 query heads and 3 key-value heads, each with head dimension 128; thus, four query heads share each key-value head. 
It uses the Llama 3 RoPE scaling rule with base frequency \(\theta=5\times10^{5}\), an original context length of 8,192, and a configured maximum context length of 131,072. RMSNorm uses \(\epsilon=10^{-5}\). 
The input embedding and output language-model head are not tied. 
The resulting non-recurrent 20-layer model contains exactly \(1{,}007{,}482{,}368\)
parameters.

The Qwen3-0.6B reference configuration contains 28 Transformer blocks with \(d_{\mathrm{model}}=1024\) and \(d_{\mathrm{ff}}=3072\). 
It uses 16 query heads and 8 key-value heads with head dimension 128. 
Consequently, its query projection has inner dimension \(16\times128=2048\), whereas its key and value projections have inner dimension \(8\times128=1024\). 
In addition to the RMSNorms surrounding the two sublayers, Qwen3-0.6B applies RMSNorm to the query and key vectors before RoPE (QK-Norm). 
It uses standard RoPE with \(\theta=10^{6}\), and a configured maximum context length of 40,960. 
RMSNorm uses \(\epsilon=10^{-6}\), and the input embedding and language-model head are tied. 
The complete 28-layer reference model contains \(596{,}049{,}920\) parameters.
Tab.~\ref{tab:backbone-configs} summarizes the principal architectural details of these two backbones.

\begin{table*}[h]
\centering
\scriptsize
\renewcommand{\arraystretch}{1.22}
\setlength{\tabcolsep}{11pt}
\setlength{\abovecaptionskip}{3pt}
\setlength{\belowcaptionskip}{5pt}
\begin{tabular}{lccccccc}
\textbf{Architecture} & \(\boldsymbol{L_{\mathrm{ref}}}\) & \textbf{Parameters} & \(\boldsymbol{|\mathcal{V}|}\) & \(\boldsymbol{d_{\mathrm{model}}}\) & \(\boldsymbol{d_{\mathrm{ff}}}\) & \textbf{Q/KV heads} & \(\boldsymbol{d_{\mathrm{head}}}\)\\
\shline
 Llama3.1-1B
 & 20 & 1,007,482,368 & 128,256 & 1,536 & 5,376 & 12/3 & 128\\
 Qwen3-0.6B
 & 28 & 596,049,920 & 151,936 & 1,024 & 3,072 & 16/8 & 128\\
\end{tabular}
\caption{\footnotesize Non-recurrent reference architectures. \(L_{\mathrm{ref}}\) denotes the number of Transformer blocks in the corresponding reference configuration. Parameter counts include token embeddings, the final RMSNorm, and the language-model head.}
\label{tab:backbone-configs}
\vspace{-3mm}
\end{table*}

\paragraph{Pretraining data and tokenization.}

All models are pretrained on the same raw FineWeb-Edu-350BT corpus~\cite{fineweb}, with a separate held-out FineWeb-Edu validation split. 
We stream the training data and shuffle examples using a buffer of 10,000 documents and random seed 42.
No data curriculum strategy is applied.

We use the tokenizer of Llama3.1-8B for the Llama3.1-1B models and the tokenizer of official Qwen3-0.6B for the Qwen3-0.6B models, yielding vocabulary sizes of 128,256 and 151,936, respectively. 
Documents are tokenized without a beginning-of-sequence token and are terminated with an end-of-sequence token.
The resulting token stream is concatenated and packed into non-overlapping sequences of length 2,048, and incomplete final sequences are discarded. 
Unless otherwise specified, the Llama3.1-1B experiments process approximately 20.15B packed tokens, while the Qwen3-0.6B experiments process approximately 12B packed tokens, which follows 1x Chinchilla optimal ratio.

\paragraph{BaseLoop and CoreLoop models.}

We describe a recurrent architecture by
\[
    p+s\times K+c,
\]
where \(p\) is the number of non-recurrent Prelude blocks, \(s\) is the number of parameterized blocks in the recurrent core, \(K\) is the number of training-time loop count of that core, and \(c\) is the number of non-recurrent Coda blocks. 
Its physical depth and training effective depth are therefore
\[
    L_{\mathrm{phys}}=p+s+c,
    \qquad
    L_{\mathrm{eff}}=p+Ks+c.
\]
The Prelude and Coda blocks have independent parameters and are each executed once, whereas the same \(s\) core blocks are reused at every recurrent iteration.

A \textit{BaseLoop} model sets \(p=c=0\), so that all physical Transformer blocks belong to the recurrent stack. 
A \textit{CoreLoop} model places one or more independently parameterized blocks before or after a smaller recurrent core. 
This construction allows us to vary where the parameterized blocks are
allocated while holding \textbf{both} \(L_{\mathrm{phys}}\) and \(L_{\mathrm{eff}}\)
fixed.
We train the Llama3.1-1B variants at \(L_{\mathrm{eff}}=20\), matching the depth of the corresponding non-recurrent reference model, and the Qwen3-0.6B variants at \(L_{\mathrm{eff}}=28\). 
Notably, although recurrent models match their reference architecture in effective depth, their parameter counts depend on the number of physical blocks 
\(L_{\mathrm{phys}}\), rather than on \(L_{\mathrm{eff}}\). 
The complete configurations of BaseLoop and CoreLoop models are  given in Tab.~\ref{tab:allocation-matrix}.

\begin{table*}[h]
\centering
\scriptsize
\renewcommand{\arraystretch}{1.22}
\setlength{\tabcolsep}{7pt}
\setlength{\abovecaptionskip}{3pt}
\setlength{\belowcaptionskip}{5pt}
\begin{tabularx}{\linewidth}{lcccc>{\raggedright\arraybackslash}X}
\textbf{Backbone} & \(\boldsymbol{L_{\mathrm{eff}}}\) & \(\boldsymbol{L_{\mathrm{phys}}}\) & \textbf{Parameters} & \textbf{BaseLoop} & \textbf{CoreLoop}\\
\shline
 Llama3.1-1B
 & 20 & 2 & 455,351,808 & \(2\times10\)
 & \(0+1\times19+1\), \(1+1\times19+0\)\\
 Llama3.1-1B
 & 20 & 4 & 516,699,648 & \(4\times5\)
 & \(0+2\times9+2\), \(1+2\times9+1\), \(2+2\times9+0\)\\
 Llama3.1-1B
 & 20 & 5 & 547,373,568 & \(5\times4\)
 & \(0+3\times6+2\), \(1+3\times6+1\), \(2+3\times6+0\)\\
 Llama3.1-1B
 & 20 & 10 & 700,743,168 & \(10\times2\)
 & \(0+5\times3+5\), \(1+5\times3+4\), \(2+5\times3+3\), \(3+5\times3+2\), \(4+5\times3+1\), \(5+5\times3+0\)\\
 Qwen3-0.6B
 & 28 & 4 & 218,507,264 & \(4\times7\)
 & \(0+2\times13+2\), \(1+2\times13+1\), \(2+2\times13+0\)\\
\end{tabularx}
\caption{\footnotesize Details of BaseLoop and CoreLoop models. An allocation is written as \(p+s\times K+c\). Architectures within a row have the same physical depth, total parameter count, and training effective depth. The reported parameter counts correspond to the unconditioned naive-loop models, without adding any input-injection or timestep-conditioning parameters.}
\label{tab:allocation-matrix}
\vspace{-1mm}
\end{table*}

At inference time, the recurrent core can instead be applied \(r\) times, resulting in inference depth of
\[
    L_{\mathrm{inference}}(r)=p+rs+c.
\]

Consequently, comparisons at different \(r\) account for the resulting operator-level compute rather than equating models solely by their raw loop counts.

\paragraph{Initial-state injection models.}

The initial-state injection experiments use a decoder-only BaseLoop architecture consisting of token embeddings, a shared stack of $K$ Transformer layers, a final RMSNorm, and a vocabulary projection.
The $K$ layers have distinct parameters, but the entire stack is reused at every recurrent iteration.
The prelude and coda Transformer depths are both zero, so $h_0$ is
directly the token embedding and the effective depth is $D=KL$.
The initial-state injection adapter is placed at the entrance to the shared stack. 
Its parameters are shared over all iterations and token positions.
Training uses next-token cross-entropy on the final recurrent output and full backpropagation through the fixed number of training iterations.

The Qwen3-0.6B corresponding BaseLoop parameter counts, before adding injection, are approximately $187.05$M, $218.51$M, $265.70$M, and $375.82$M.
Thus, Qwen3-0.6B identifies the backbone configuration; the recurrent models have fewer distinct parameters because of layer sharing.
The additional Llama experiments use the Llama3.1-1B backbone. 
We use $K\times L_{\mathrm{train}}\in
\{2\times10,4\times5,5\times4,10\times2\}$, all with effective
training depth $20$.
Their BaseLoop parameter counts are approximately
$455.35$M, $516.70$M, $547.37$M, and $700.74$M, respectively.
All four initial-state injection variants are evaluated for every
Qwen and Llama configuration, alongside the corresponding BaseLoop
without injection.
The non-loop references have $28$ layers for Qwen3-0.6B and $20$ layers for Llama3.1-1B.

For Scalar injection, we initialize $\alpha=0$.
For Channel-wise injection, we initialize $a=\mathbf{1}$ and
$b=\mathbf{0}$; for Residual Channel-wise injection, we initialize
$\delta_a=b=\mathbf{0}$.
Dense injection implements a bias-free projection from the concatenated state and initial representation, with weight
$[W_h\;W_0]\in\mathbb{R}^{d\times2d}$ initialized to $[I_d\;0]$.
These initializations make the injection maps initially equal to the identity on the current state.
The additional parameter counts are $1$, $2d$, $2d$, and $2d^2$,
respectively.
All coefficients are unconstrained learned parameters.
The two channel-wise variants have the same function class but
different parameterizations.
With the configured weight decay, regularizing $a$ favors a state scale of zero, whereas regularizing $\delta_a$ favors a state scale of one.

\paragraph{History-state injection models.}
All reported history-state experiments use the Qwen3-0.6B BaseLoop
$4\times7$ architecture and the same $12$B-token training recipe as
the corresponding initial-state injection models.
We instantiate Eq.~\ref{eq:history_state_injection} with the
scalar form of the lag-specific operators,
$\mathcal{B}_j(x)=\beta_j x$ with $\beta_j\in\mathbb{R}$,
so that each recurrent transition becomes
\begin{equation}
    h_{\ell+1}
    =F_\theta\biggl(
    h_\ell+\sum_{j=1}^{m_\ell}\beta_j\,(h_{\ell-j}-h_\ell)
    \biggr),
    \qquad
    m_\ell=\min\{w,\max(\ell-1,0)\}.
    \label{eq:history_scalar}
\end{equation}

\textit{History window.}
For the transition $h_\ell\to h_{\ell+1}$, the history window contains
the $m_\ell$ most recent completed recurrent states
$h_{\ell-1},\ldots,h_{\ell-m_\ell}$, indexed by relative lag $j$:
the state at lag $j$, $h_{\ell-j}$, is weighted by $\beta_j$.
The window excludes the current state $h_\ell$, which serves as the
reference point of every difference, and the embedding state $h_0$,
which is reserved for initial-state injection.
Consequently, $m_0=m_1=0$, and the first two transitions reduce to
plain BaseLoop recurrence, $h_1=F_\theta(h_0)$ and $h_2=F_\theta(h_1)$.
The first history-dependent transition is
\begin{equation}
    h_3=F_\theta\bigl(h_2+\beta_1(h_1-h_2)\bigr).
\end{equation}
Once more than $w$ completed states are available ($\ell-1>w$),
only the $w$ most recent ones are retained.
Buffered states are not detached, so gradients propagate through
them during training.

\textit{Parameterization and initialization.}
The coefficients $\beta_1,\ldots,\beta_w$ are unconstrained,
initialized to zero, and depend only on the relative lag $j$;
they are shared across all recurrent iterations and token positions.
History-only injection therefore adds exactly $w$ trainable scalars
to BaseLoop and recovers BaseLoop recurrence exactly at initialization.

\textit{Combined variant.}
When history-state injection is combined with initial-state
injection, each transition becomes
\begin{equation}
    h_{\ell+1}
    =F_\theta\biggl(
    h_\ell+\alpha h_0+
    \sum_{j=1}^{m_\ell}\beta_j\,(h_{\ell-j}-h_\ell)
    \biggr),
\end{equation}
with $\alpha=0$ and $\beta_j=0$ for all $j$ at initialization.
The differences are always taken with respect to the recurrent state
$h_\ell$, not the injected input $h_\ell+\alpha h_0$,
so the two branches enter additively and do not interact.

\paragraph{Timestep-conditioning models.}

Let $T$ denote the number of steps in the conditioning grid.
At recurrent iteration $\ell$, we set
$t_\ell=\ell/T$ and $\Delta t=1/T$.
The conditioning vector is
$\psi_\ell=\psi(t_\ell,\Delta t)\in\mathbb{R}^{8}$, where
\begin{equation}
\begin{aligned}
\psi(t,\Delta t)=\bigl(
&t,\Delta t,t^2,(\Delta t)^2,t\Delta t,\\
&\sin(\pi t),\cos(\pi t)-1,\sin(2\pi t)
\bigr)^\top.
\end{aligned}
\label{eq:timestep_features}
\end{equation}

For \textit{Loop Gating}, a learned vector
$q\in\mathbb{R}^{8}$ determines the update:
\begin{equation}
\begin{aligned}
g_\ell &= 1+q^\top\psi_\ell,\\
h_{\ell+1}
&=h_\ell+g_\ell\bigl(F_\theta(h_\ell)-h_\ell\bigr).
\end{aligned}
\label{eq:loop_level_timestep_gating}
\end{equation}

For the other two variants, let
$k\in\{1,\ldots,K\}$ index a layer and
$b\in\{\mathrm{attn},\mathrm{mlp}\}$ index its residual branch.
Writing $\mathcal{B}_k^b$ for the branch operation and $N_k^b$
for its preceding RMSNorm, the branch update is
\begin{equation}
u^+
=
u+g_{\ell,k}^{b}\odot
\mathcal{B}_k^b\!\left(
(\mathbf{1}+s_{\ell,k}^{b})\odot N_k^b(u)
\right).
\label{eq:branch_timestep_modulation}
\end{equation}
Each layer applies this update first to the attention branch
and then to the MLP branch, with $u$ denoting the current
branch input.
The backbone RMSNorm parameters are retained.

\textit{Branch Gating} uses
\begin{equation}
g_{\ell,k}^{b}
=
1+(q_k^b)^\top\psi_\ell,
\qquad
s_{\ell,k}^{b}=\mathbf{0},
\label{eq:branch_level_timestep_gating}
\end{equation}
where $q_k^b\in\mathbb{R}^{8}$ and the scalar gate is
broadcast over hidden channels.
\textit{AdaLN} instead uses
\begin{equation}
\begin{bmatrix}
g_{\ell,k}^{b}-\mathbf{1}\\
s_{\ell,k}^{b}
\end{bmatrix}
=
W_k^b\psi_\ell,
\qquad
W_k^b\in\mathbb{R}^{2d\times8}.
\label{eq:adaln_timestep_parameters}
\end{equation}
Thus, each layer generates four $d$-dimensional vectors,
with no timestep-dependent additive shift.
These linear maps implement our parameterization of the
AdaLN modulation.

All conditioning weights are learned jointly with the backbone
and shared across token positions and recurrent iterations.
The weights indexed by $k$ and $b$ are distinct across layers
and branches.
We initialize $q$, $q_k^b$, and $W_k^b$ to zero, so every gate
initially equals one and every additional normalization scale
equals zero, recovering the unconditioned BaseLoop computation.
The gates and scales are unconstrained.
Loop Gating, Branch Gating, and AdaLN add $8$, $16K$, and $32Kd$ trainable parameters, respectively.

During training, $T=L_{\mathrm{train}}$.
For an inference budget of $L_{\mathrm{infer}}$ iterations,
we consider two conditioning grids.
\textit{Prefix} inference retains $T=L_{\mathrm{train}}$
and executes the first $L_{\mathrm{infer}}$ steps,
requiring $L_{\mathrm{infer}}\leq L_{\mathrm{train}}$.
\textit{Rescaled} inference sets $T=L_{\mathrm{infer}}$,
so that
$t_\ell=\ell/L_{\mathrm{infer}}$ and
$\Delta t=1/L_{\mathrm{infer}}$;
this supports budgets both below and above the training budget.
In both cases, $\ell=0,\ldots,L_{\mathrm{infer}}-1$.
The step size enters through the conditioning features;
the updates above have no additional $\Delta t$ multiplier.

\subsection{Training}
\label{app:training}

\paragraph{Recurrent backpropagation.}

All LoopLM models are trained from scratch using the standard autoregressive
language-modeling objective. 
Given a token sequence \((x_1,\ldots,x_T)\), we minimize the mean next-token cross-entropy
\[
    \mathcal{L}_{\mathrm{LM}}
    =
    -\frac{1}{T-1}
    \sum_{t=1}^{T-1}
    \log p_{\theta}(x_{t+1}\mid x_{\leq t}),
\]
where the loss is averaged over all non-masked target tokens. 

During training, each recurrent core is executed using the fixed loop count \(K\) specified by its architecture. 
Gradients are propagated through all \(K\) loops of the recurrent core using full backpropagation through time. 
Consequently, the gradient of each shared core block aggregates its contributions from every recurrent iteration. 
No recurrent iteration is detached from the computation graph.

\paragraph{Optimizer.}

We optimize all models using Muon~\cite{muon,dmuon}. 
Specifically, Muon is used for the matrix-valued attention and feed-forward projection weights. 
Token embeddings, the language-model head, normalization parameters, and other vector- or scalar-valued parameters are optimized using AdamW~\cite{adamw}.
For Muon parameter group, we use momentum \(0.95\), Nesterov momentum, and
five-step Newton-Schulz iterations per optimizer update. 
For AdamW parameter
group, we use
\[
    (\beta_1,\beta_2)=(0.9,0.95),
    \qquad
    \epsilon_{\mathrm{AdamW}}=10^{-8}.
\]
A decoupled weight decay of \(0.1\) is applied. The configured peak learning
rates are
\[
    \eta_{\max}^{\mathrm{Llama}}=1.33847\times10^{-3},
    \qquad
    \eta_{\max}^{\mathrm{Qwen}}=1.80111\times10^{-3}.
\]

\paragraph{Learning-rate scheduler.}

All LoopLM experiments use a linear warmup-stable-decay (WSD) scheduler.
The learning rate increases linearly from zero to \(\eta_{\max}\) during the first \(5\%\) of optimizer updates, remains at \(\eta_{\max}\) for the next \(85\%\), and then decreases linearly during the final \(10\%\). 
The terminal learning rate is set as \(0.1\eta_{\max}\). 
More precisely, for warmup, stable, and decay lengths \(T_{\mathrm{w}}\),
\(T_{\mathrm{s}}\), and \(T_{\mathrm{d}}\), respectively, the learning-rate
multiplier is
\[
    \lambda(t)=
    \begin{cases}
        t/T_{\mathrm{w}},
        & 0\leq t\leq T_{\mathrm{w}}, \\[2mm]
        1,
        & T_{\mathrm{w}}<t\leq T_{\mathrm{w}}+T_{\mathrm{s}}, \\[2mm]
        1-0.9
        \dfrac{t-T_{\mathrm{w}}-T_{\mathrm{s}}}{T_{\mathrm{d}}},
        & T_{\mathrm{w}}+T_{\mathrm{s}}<t
          \leq T_{\mathrm{w}}+T_{\mathrm{s}}+T_{\mathrm{d}}.
    \end{cases}
\]
The learning rate at update \(t\) is \(\eta(t)=\eta_{\max}\lambda(t)\). 
Exact schedule lengths and batch sizes are reported in Tab.~\ref{tab:training-recipes}.

\begin{table*}[t]
\centering
\scriptsize
\renewcommand{\arraystretch}{1.22}
\setlength{\tabcolsep}{7pt}
\setlength{\abovecaptionskip}{3pt}
\setlength{\belowcaptionskip}{5pt}
\begin{tabular}{lccccc}
\textbf{Models} & \(\boldsymbol{\eta_{\max}}\) & \textbf{Global batch} & \textbf{Steps} & \textbf{W/S/D updates} & \textbf{Tokens}\\
\shline
 Llama3.1-1B BaseLoop and CoreLoop variants
 & \(1.33847\times10^{-3}\) & 1,824 & 5,394 & 270/4,585/539 & 20.15B\\
 Qwen3-0.6B BaseLoop and CoreLoop variants
 & \(1.80111\times10^{-3}\) & 2,048 & 2,860 & 143/2,431/286 & 12.00B\\
\end{tabular}
\caption{\footnotesize Training settings for LoopLM models. Global batch size is measured in packed sequences of length 2,048. W/S/D gives the exact numbers of warmup, stable, and decay updates. Token counts are computed before the one-token shift used by the next-token objective.}
\label{tab:training-recipes}
\vspace{-3mm}
\end{table*}

\paragraph{Gradient accumulation and clipping.}

The Llama3.1-1B runs with global batch size 1,824 use a per-device micro-batch size
of 38 and accumulate gradients over six micro-batches. 
The Qwen3-0.6B runs use a per-device micro-batch size of 32 and accumulate gradients over eight micro-batches.
Losses are divided by the number of accumulation steps before backpropagation. 
After the accumulated gradients have been synchronized and before each optimizer update, we clip the global \(\ell_2\) norm of all model gradients to \(1.0\).

\paragraph{Precision, initialization, and reproducibility.}

Training uses bfloat16 model parameters and bfloat16 mixed-precision computation. Linear and embedding weights are initialized independently from a zero-mean Gaussian distribution with standard deviation \(0.02\). 
All experiments use random seed 42.

\subsection{Evaluation}
\label{app:evaluation}

All benchmarks are evaluated in zero-shot mode with \texttt{lm-evaluation-harness}~\cite{eval-harness}. 
Notably, we use length-normalized accuracy for multiple-choice tasks whose options differ in length, and plain accuracy for tasks with fixed-form options. 
Every task is evaluated at each inference loop count $r$ independently, so a full evaluation sweep yields one accuracy value per task per $r$.

\paragraph{Task groups.}

The \textit{Knowledge} group (Tab.~\ref{tab:knowledge-tasks}) contains three tasks and the \textit{Reasoning} group (Tab.~\ref{tab:reasoning-tasks}) eight tasks. 
Group scores are unweighted averages,
\begin{equation}
  S_{\mathrm{Know}} = \tfrac{1}{3}\!\!\sum_{i\in\mathcal{T}_{\mathrm{Know}}}\!\! a_i ,
  \qquad
  S_{\mathrm{Reas}} = \tfrac{1}{8}\!\!\sum_{i\in\mathcal{T}_{\mathrm{Reas}}}\!\! a_i ,
\end{equation}
where $a_i$ is the headline metric of task $i$. 
Equal weighting is deliberate: it prevents a group score from being dominated by whichever benchmark happens to have the largest dynamic range at this model scale, at the cost of giving each task equal influence regardless of test-set size.
Notably, we assign PIQA to the knowledge group even though it is commonly described as physical commonsense reasoning, because its instances are resolved via stored knowledge of object affordance and material properties.

\begin{table}[h]
\centering
\scriptsize
\renewcommand{\arraystretch}{1.22}
\setlength{\tabcolsep}{10pt}
\setlength{\abovecaptionskip}{3pt}
\setlength{\belowcaptionskip}{5pt}
\begin{tabular}{ll}
\textbf{Task} & \textbf{Brief Description}\\
\shline
 \textsc{SciQ}~\cite{sciq} & Science facts; locating evidence in a support passage\\
 \textsc{ARC-Easy}~\cite{arc} & Elementary science knowledge and simple causal attribution\\
 \textsc{PIQA}~\cite{piqa} & Object affordances, material properties, outcomes of actions\\
\end{tabular}
\caption{\footnotesize Three tasks in the \textit{Knowledge} group.}
\label{tab:knowledge-tasks}
\vspace{-3mm}
\end{table}

\begin{table}[h]
\centering
\scriptsize
\renewcommand{\arraystretch}{1.22}
\setlength{\tabcolsep}{10pt}
\setlength{\abovecaptionskip}{3pt}
\setlength{\belowcaptionskip}{5pt}
\begin{tabular}{ll}
\textbf{Task} & \textbf{Brief Description}\\
\shline
 \textsc{ARC-Challenge}~\cite{arc} & Multi-step inference over harder science items\\
 \textsc{WinoGrande}~\cite{winogrande} & Coreference resolution requiring commonsense\\
 \textsc{OpenBookQA}~\cite{openbookqa} & Combining a retrieved science fact with additional knowledge\\
 \textsc{HellaSwag}~\cite{hellaswag} & Plausible event continuation; temporal and causal structure\\
 \textsc{CommonsenseQA}~\cite{commonsenseqa} & ConceptNet-style relational reasoning over use, location\\
 \textsc{ProofWriter}~\cite{proofwriter} & Entailment under natural-language facts and rules\\
 \textsc{CLUTRR}~\cite{clutrr} & Kinship inference along a relation chain in a short story\\
 \textsc{BBH}~\cite{bbh} & Logical deduction, temporal ordering, object-state tracking\\
\end{tabular}
\caption{\footnotesize Eight tasks in the \textit{Reasoning} group. The last three tasks are the ones whose difficulty is controlled by an explicit compositional parameter.}
\label{tab:reasoning-tasks}
\vspace{-3mm}
\end{table}

\begin{table*}[tb]
\centering
\scriptsize
\renewcommand{\arraystretch}{1.22}
\setlength{\tabcolsep}{7pt}
\setlength{\abovecaptionskip}{3pt}
\setlength{\belowcaptionskip}{5pt}
\begin{tabularx}{\linewidth}{l>{\raggedright\arraybackslash}X}
\textbf{Subtask} & \textbf{Brief Description}\\
\shline
 \path{disambiguation_qa} & Resolves pronoun antecedents and identifies cases that remain genuinely ambiguous.\\
 \path{salient_translation_error_detection} & Classifies salient semantic errors in German-to-English translations, such as altered entities, numbers, negation, or omitted content.\\
 \path{reasoning_about_colored_objects} & Answers attribute, spatial-relation, and counting questions about colored objects described in natural language.\\
 \path{causal_judgement} & Determines commonsense causal attribution and whether an action or outcome was intentional.\\
 \path{date_understanding} & Infers calendar dates from relative temporal expressions and performs date arithmetic.\\
 \path{hyperbaton} & Selects the sentence exhibiting the grammatically natural ordering of English adjectives.\\
 \path{logical_deduction_three_objects} & Infers the ordering of three objects from a set of logically consistent relational constraints.\\
 \path{logical_deduction_seven_objects} & Infers the ordering of seven objects from relational constraints, requiring a larger reasoning state.\\
 \path{penguins_in_a_table} & Performs lookup, comparison, counting, sorting, and update operations over a semi-structured table.\\
 \path{snarks} & Identifies which of two statements is sarcastic, testing pragmatic and contextual language understanding.\\
 \path{temporal_sequences} & Finds a feasible time interval by reasoning over schedules, event durations, and temporal constraints.\\
 \path{tracking_shuffled_objects_three_objects} & Tracks a sequence of pairwise swaps among three entities to determine the final object assignment.\\
 \path{tracking_shuffled_objects_five_objects} & Tracks a sequence of pairwise swaps among five entities, increasing the required state-tracking capacity.\\
 \path{tracking_shuffled_objects_seven_objects} & Tracks a sequence of pairwise swaps among seven entities, providing the most demanding state-tracking variant.\\
\end{tabularx}
\caption{\footnotesize The selected 14 \textsc{BBH} subtasks and their evaluated capabilities.}
\label{tab:bbh_subtasks}
\vspace{-3mm}
\end{table*}

\paragraph{Selected BBH subtasks.}

These 14 subtasks cover complementary dimensions of reasoning, as shown in Tab.~\ref{tab:bbh_subtasks}.
\texttt{disambiguation\_qa} tests pronoun resolution and the recognition of
genuine referential ambiguity;
\texttt{salient\_translation\_error\_detection} requires classifying semantic
errors in German-to-English translations;
\texttt{reasoning\_about\_colored\_objects} evaluates attribute retrieval,
spatial relations, and counting over described objects.
\texttt{causal\_judgement} assesses commonsense causal attribution and
intentionality, whereas \texttt{date\_understanding} tests calendar arithmetic.
\texttt{hyperbaton} evaluates knowledge of English adjective ordering.
The two \texttt{logical\_deduction} variants require recovering an
ordering from relational constraints.
\texttt{penguins\_in\_a\_table} requires structured table lookup,
comparison, counting, and sorting.
\texttt{snarks} probes pragmatic reasoning through sarcasm detection.
\texttt{temporal\_sequences} tests temporal-constraint reasoning.
Finally, the three \texttt{tracking\_shuffled\_objects} variants require maintaining
entity--object assignments through a sequence of swaps. The variants with
different numbers of objects provide a controlled measure of how performance
changes as the amount of relational state to be tracked increases.
Notably, the remaining \textsc{BBH} subtasks were excluded because they proved excessively challenging for models at the evaluated scale. 
Across model variants, performance on most of these tasks fluctuated around their
task-specific random-guessing baselines and showed no consistent separation
between models. 
Consequently, they provided little discriminative signal for meaningful model comparison and were not included in the aggregate score.

\subsection{Geometry Probing}
\label{app:probing}

\textbf{Execution and sampling.}
We probe frozen checkpoints in evaluation mode on held-out validation set.
The reported Qwen3-0.6b and Llama3.1-1B comparisons each use 49 packed sequences
of maximum length 2,048, with exactly 100,000 selected next-token
prediction positions. 
Within each model family, the tokenized examples and selection masks are identical across configurations and inference depths. 
Let $\mathcal{T}$ contain these selected positions, $N$ be the number of nonempty sequences, and $p_n$ be the last selected prediction
position in sequence $n$. 
\textit{Angular distance} and \textit{Relative update norm} use one position $p_n$ per sequence; \textit{Variance} and \textit{update-response} scores use all positions in $\mathcal{T}$. 
The selection follows the next-token objective, including the partially selected final sequence at the token budget boundary.

Let $h_{\ell,p}^{\mathrm{in}},h_{\ell,p}^{\mathrm{out}}\in\mathbb{R}^{m}$
denote the actual input and output of the block executed at effective
layer position $\ell$, for token position $p$. Repeated executions of a
shared block have distinct effective positions, even though they use the
same parameters. For prelude depth $a$, shared depth $s$, coda depth $b$,
and $L$ inference loops, the evaluated layouts have effective depth
$D=a+sL+b$. We also record $x_p^{(t)}$, the recurrent state after $t$
complete loops, for $t=0,\ldots,L$; $x_p^{(0)}$ is the state entering the
first loop after any prelude. A complete loop includes state injection,
the shared blocks, and any loop-level mixing or gating. Block probes
use each block's own boundaries, whereas loop probes include these
additional operations. In particular, injection can make a block's
input differ from the preceding block's output. Recurrent-state probes
exclude the coda and final output normalization.

\paragraph{Angular distance.}
For two state vectors $u,v$, we compute normalized angular distance
\begin{align}
    c_{\epsilon}(u,v)
    &=\operatorname{clip}_{[-1,1]}\!\left(
       \frac{u^{\top}v}
       {\max(\|u\|_2\|v\|_2,\epsilon)}\right),\\
    a(u,v)&=\frac{\arccos c_{\epsilon}(u,v)}{\pi},
    \qquad \epsilon=10^{-12}.
    \label{eq:probing_angle}
\end{align}
The block and loop measurements are
\begin{align}
    A_{\ell}
    &=\frac{1}{N}\sum_{n=1}^{N}
      a(h_{\ell,p_n}^{\mathrm{in}},h_{\ell,p_n}^{\mathrm{out}}),\\
    A_{\mathrm{loop}}(t)
    &=\frac{1}{N}\sum_{n=1}^{N}
      a(x_{p_n}^{(t)},x_{p_n}^{(t+1)}),
      \quad 0\leq t<L.
\end{align}
For nondegenerate vectors, values 0, $1/2$, and 1 indicate aligned,
orthogonal, and opposite directions, respectively. We average the
individual angular distances, rather than applying $\arccos$ to an
averaged cosine similarity. Clipping prevents numerical overshoots;
the denominator floor assigns distance $1/2$ when either vector is zero.
The metric measures directional change and is insensitive to positive
rescaling away from the numerical floor.

\paragraph{Relative update norm.}
We measure the update magnitude relative to the incoming state:
\begin{align}
    \rho_{\ell}
    &=\frac{1}{N}\sum_{n=1}^{N}
      \frac{\|h_{\ell,p_n}^{\mathrm{out}}
               -h_{\ell,p_n}^{\mathrm{in}}\|_2}
           {\max(\|h_{\ell,p_n}^{\mathrm{in}}\|_2,\epsilon)},\\
    \rho_{\mathrm{loop}}(t)
    &=\frac{1}{N}\sum_{n=1}^{N}
      \frac{\|x_{p_n}^{(t+1)}-x_{p_n}^{(t)}\|_2}
           {\max(\|x_{p_n}^{(t)}\|_2,\epsilon)}.
    \label{eq:probing_relative_update}
\end{align}
These are means of per-sequence ratios, not ratios of mean norms.
A value of 0.1 corresponds to a mean relative update magnitude of 10\%.
The metric can exceed one. Angular distance and relative update norm
capture different changes: multiplying a nonzero state by a positive
scalar leaves its angle unchanged but can yield a substantial relative
update. Small loop updates indicate little movement under the measured
iteration, without establishing convergence or task usefulness.

\paragraph{State variance and plot normalization.}
For a state vector $h_p\in\mathbb{R}^{m}$, we first compute variance across
its hidden coordinates, using correction one, and then average over tokens:
\begin{align}
    \bar h_p&=\frac{1}{m}\sum_{k=1}^{m}h_{p,k},\\
    V(h)&=\frac{1}{|\mathcal{T}|}
      \sum_{p\in\mathcal{T}}\frac{1}{m-1}
      \sum_{k=1}^{m}(h_{p,k}-\bar h_p)^2.
    \label{eq:probing_variance}
\end{align}
The block-output variance is $V(h_{\ell}^{\mathrm{out}})$, and the
recurrent-state variance is $V(x^{(t)})$; block-input variances are also
recorded. This measures within-token feature dispersion, rather than
variance across examples or the rank of a representation covariance.
All selected tokens receive equal weight.

The BaseLoop/CoreLoop trajectory figure normalizes each model by its own
state variance at the training loop count $L_{\mathrm{tr}}$:
\begin{equation}
    \widetilde V(t)=
    \frac{V(x^{(t)})}{V(x^{(L_{\mathrm{tr}})})}.
    \label{eq:probing_variance_normalization}
\end{equation}
Thus, $\widetilde V(L_{\mathrm{tr}})=1$; a value above one indicates
growth relative to that model's state at the training horizon.
The block-angle horizontal axis uses effective depth $\ell$.
Loop-angle and loop-update transitions are located at
$(t+1)/L_{\mathrm{tr}}$, whereas state variances are located at
$t/L_{\mathrm{tr}}$ and include $t=0$. The reference at one marks the
training loop count. Logarithmic axes change the display scale only.

To locate amplification within a loop, let $z^{(t)}$ be the input to
its first shared block, after any state injection, and $y^{(t)}$ the
output of its last shared block, before loop-level mixing. Define
\begin{align}
    g_{\mathrm{inj}}(t)&=\frac{V(z^{(t)})}{V(x^{(t)})},&
    g_{\mathrm{core}}(t)&=\frac{V(y^{(t)})}{V(z^{(t)})},\\
    g_{\mathrm{mix}}(t)&=\frac{V(x^{(t+1)})}{V(y^{(t)})}.
\end{align}
For positive variances, these factors telescope exactly:
\begin{equation}
    \frac{V(x^{(L)})}{V(x^{(L_{\mathrm{tr}})})}
    =\prod_{t=L_{\mathrm{tr}}}^{L-1}
      g_{\mathrm{inj}}(t)g_{\mathrm{core}}(t)g_{\mathrm{mix}}(t).
\end{equation}
This accounting identifies the stage at which variance grows. The
factors are ratios of token-averaged variances, not additive contribution
fractions or means of per-token variance ratios.

\paragraph{Computational interaction.}
For effective positions $i<j$, we run the same examples twice: once
normally and once replacing the output of block occurrence $i$ by its
input. Only that occurrence is skipped; other executions of the shared
parameters remain active, and the remaining computation proceeds normally.
At target position $j$, define
\begin{align}
    u_{j,p}&=h_{j,p}^{\mathrm{out}}-h_{j,p}^{\mathrm{in}},\\
    u_{j,p}^{(-i)}
    &=h_{j,p}^{\mathrm{out},(-i)}-h_{j,p}^{\mathrm{in},(-i)}.
\end{align}
Each update uses the input and output from its own execution. The
response numerator and baseline update magnitude are
\begin{align}
    N_{i\to j}&=\frac{1}{|\mathcal{T}|}
      \sum_{p\in\mathcal{T}}\|u_{j,p}^{(-i)}-u_{j,p}\|_2,\\
    B_j&=\frac{1}{|\mathcal{T}|}
      \sum_{p\in\mathcal{T}}\|u_{j,p}\|_2,
\end{align}
and the relative response is
\begin{equation}
    C_{i\to j}=\frac{N_{i\to j}}{B_j}.
    \label{eq:probing_response}
\end{equation}
Unlike $\rho$, $C$ is a ratio of token-averaged norms. For example,
$C=0.1$ means that the mean change in the downstream update vector is
one tenth of its baseline mean norm. It measures the change in the
update, including its direction, rather than only the change in update
magnitude or the difference between output states.
The implementation records zero when $B_j\leq10^{-8}$; all pairs in
the main response figure exceed this threshold. This fallback should
not be interpreted as a measured absence of influence.

\paragraph{Aggregation by loop lag.}
For Qwen BaseLoop $4\times7$, the loop containing effective position $j$
is $r(j)=1+\lfloor(j-1)/4\rfloor$. We include only pairs whose source and
target both lie beyond the seven training loops. For lag $d$, define
\begin{align}
    \mathcal{P}_d
    &=\{(i,j):1\leq i<j\leq4L,\ r(i)>7,\ r(j)>7,\nonumber\\
    &\hspace{37mm}r(j)-r(i)=d\},\\
    C(d)&=\frac{1}{|\mathcal{P}_d|}
      \sum_{(i,j)\in\mathcal{P}_d}C_{i\to j}.
    \label{eq:probing_lag_response}
\end{align}
Here $d=0$ compares different blocks in the same loop, $d=1$ compares
adjacent loops, and larger values compare more widely separated loops.
The pair counts are $|\mathcal{P}_0|=6(L-7)$ and
$|\mathcal{P}_d|=16(L-7-d)$ for $1\leq d\leq L-8$.
Accordingly, the full lag ranges are 0--6 at $D=56$ ($L=14$) and
0--13 at $D=84$ ($L=21$). Panels (b,c) plot these $C(d)$ values.

For a lag set $\mathcal{D}$, the numerical summaries in the main text use
\begin{equation}
    C_{\mathcal{D}}=
    \frac{\sum_{d\in\mathcal{D}}|\mathcal{P}_d|C(d)}
         {\sum_{d\in\mathcal{D}}|\mathcal{P}_d|}.
\end{equation}
Thus, individual block pairs receive equal weight. The common cross-loop
range is $\mathcal{D}=\{1,\ldots,6\}$; the triple comparison also uses
$\mathcal{D}=\{2,\ldots,6\}$. Percent differences between configurations
$A$ and $B$ are $100(C_{\mathcal{D}}^A/C_{\mathcal{D}}^B-1)$.
Panel (d) instead plots the pointwise ratio
\begin{equation}
    R_T(d)=\frac{C_{\mathrm{I+H+T}}(d)}{C_{\mathrm{H+T}}(d)},
    \qquad T\in\{\mathrm{LG},\mathrm{BG}\},
\end{equation}
matching history parameterization, window size, gate type, and inference
depth. Ratios below one indicate a smaller mean response in the triple.
These are ratios of configuration-level means, not means of matched
pairwise ratios. Different lags contain different absolute positions and
numbers of pairs; the curves do not follow one fixed perturbation as it
travels through the network.

\newpage
\section{Full Experimental Results}

\subsection{Optimizer Experiment}
\label{app:muon}

\paragraph{Setup.}

We conducted a pilot study to determine the optimizer used in our main pre-training experiments. 
We compare AdamW and Muon under the same pre-training setup across six architectures, including standard non-recurrent Transformers with 12 unique layers (\texttt{nonloop 12$\times$1}) and 3 unique layers (\texttt{nonloop 3$\times$1}), as well as four LoopLM architectures: BaseLoop (\texttt{3$\times$4}), LoopFormer (\texttt{3$\times$4})~\cite{loopformer}, Ouro (\texttt{3$\times$4})~\cite{ouro}, and Huginn (\texttt{1+2$\times$5+1})~\cite{huginn}. 
Here, the notation describes the arrangement of unique and recurrent layers for each architecture. 
After pre-training, we evaluate all successfully trained models on ten commonly used language understanding and commonsense reasoning benchmarks: ARC-Easy (ARC-E), ARC-Challenge (ARC-C)~\cite{arc}, SciQ~\cite{sciq}, MMLU~\cite{mmlu}, HellaSwag (HELLA)~\cite{hellaswag}, OpenBookQA (OBQA)~\cite{openbookqa}, PIQA~\cite{piqa}, RACE~\cite{race}, WinoGrande (WINO)~\cite{winogrande}, and CommonsenseQA (CSQA)~\cite{commonsenseqa}. 
The average score across these benchmarks is reported as the overall performance.

\begin{table*}[h]
\centering
\scriptsize
\renewcommand{\arraystretch}{1.22}
\setlength{\tabcolsep}{4.8pt}
\setlength{\abovecaptionskip}{3pt}
\setlength{\belowcaptionskip}{5pt}
\begin{tabular}{lccccccccccc}
\textbf{Model} & \textbf{ARC-E} & \textbf{ARC-C} & \textbf{SciQ} & \textbf{MMLU} & \textbf{HELLA} & \textbf{OBQA} & \textbf{PIQA} & \textbf{RACE} & \textbf{WINO} & \textbf{CSQA} & \textbf{Avg.}\\
\shline
 \multicolumn{12}{l}{\textit{AdamW}}\\
 nonloop (12$\times$1) & 59.30 & 25.68 & 81.50 & 26.06 & 33.64 & 22.00 & 67.52 & 31.67 & 51.46 & 20.64 & 41.95\\
 nonloop (3$\times$1) & 53.41 & 20.90 & 74.30 & 23.10 & 29.04 & 18.40 & 63.38 & 27.56 & 50.99 & 20.31 & 38.14\\
 BaseLoop (3$\times$4) & 56.23 & 23.81 & 77.00 & 23.00 & 31.05 & 21.40 & 64.20 & 30.33 & 53.28 & 19.49 & 39.98\\
 LoopFormer (3$\times$4) & 56.73 & 23.89 & 77.90 & 24.59 & 30.97 & 20.00 & 64.64 & 29.28 & 51.62 & 19.98 & 39.96\\
 Ouro (3$\times$4) & 55.85 & 22.87 & 77.90 & 23.03 & 30.74 & 18.00 & 63.82 & 29.95 & 52.49 & 19.57 & 39.42\\
 Huginn (1+2$\times$5+1) & \multicolumn{11}{c}{\textit{Failed}}\\
\hline
 \multicolumn{12}{l}{\textit{Muon}}\\
 nonloop (12$\times$1) & 59.76 & 25.43 & 81.20 & 24.90 & 34.07 & 22.80 & 67.90 & 32.06 & 51.93 & 20.31 & \textbf{42.04}\\
 nonloop (3$\times$1) & 53.41 & 22.61 & 74.00 & 22.93 & 29.50 & 18.80 & 64.98 & 26.99 & 50.12 & 19.57 & \textbf{38.29}\\
 BaseLoop (3$\times$4) & 57.41 & 26.19 & 78.50 & 25.30 & 31.48 & 20.80 & 65.18 & 30.62 & 52.88 & 19.08 & \textbf{40.74}\\
 LoopFormer (3$\times$4) & 56.14 & 23.89 & 80.90 & 23.22 & 31.68 & 17.60 & 65.18 & 31.10 & 50.59 & 20.15 & \textbf{40.05}\\
 Ouro (3$\times$4) & 56.99 & 25.09 & 77.60 & 23.81 & 31.72 & 19.80 & 63.93 & 30.33 & 50.51 & 19.74 & \textbf{39.95}\\
 Huginn (1+2$\times$5+1) & 58.16 & 24.83 & 80.70 & 25.22 & 33.19 & 22.60 & 66.10 & 32.34 & 53.59 & 20.48 & \textbf{41.72}\\
\end{tabular}
\caption{\footnotesize Pilot study comparing AdamW and Muon for pre-training different non-recurrent and recurrent architectures. Results are reported on ten downstream evaluation benchmarks together with their average. \textit{Failed} indicates that the corresponding pre-training run did not successfully converge.}
\label{tab:optimizer_results}
\vspace{-3mm}
\end{table*}

\paragraph{Results.}

As shown in Tab.~\ref{tab:optimizer_results}, Muon consistently provides stronger overall performance than AdamW across all architectures successfully trained with both optimizers. 
In particular, the average score improves from 41.95 to 42.04 for the 12-layer non-recurrent Transformer, from 38.14 to 38.29 for the 3-layer non-recurrent Transformer, from 39.98 to 40.74 for BaseLoop, from 39.96 to 40.05 for LoopFormer, and from 39.42 to 39.95 for Ouro. 
More importantly, the Huginn \texttt{1+2$\times$5+1} configuration fails to train with AdamW, whereas Muon successfully trains the model and achieves an average score of 41.72. 
Based on both its consistently better downstream performance and  training stability, we therefore adopt Muon as the default optimizer for all main experiments.

\subsection{BaseLoop and CoreLoop}
\label{app:baseloop_coreloop_results}

\begin{figure}[t]
    \centering
    \setlength{\abovecaptionskip}{3pt}
    \setlength{\belowcaptionskip}{3pt}
    \includegraphics[width=\linewidth]{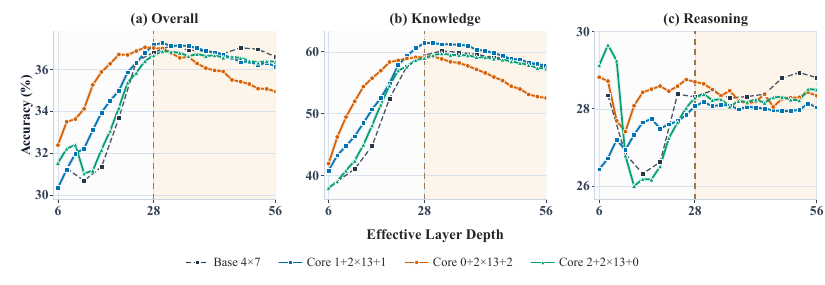}
    \caption{Average results of BaseLoop and CoreLoop models from Qwen3-0.6B with physical layers of 4 on \textit{Overall}, \textit{Knowledge}, and \textit{Reasoning} benchmarks. The dashed vertical line marks the effective training depth of 28 layers. Results at or below this depth correspond to evaluation within the training depth budget, while the yellow-shaded region denotes extrapolation beyond it.
    }
    \label{fig:app_baseloop_coreloop_qwen}
\end{figure}

Fig.~\ref{fig:app_baseloop_coreloop_qwen} shows that the preferred
CoreLoop layout on Qwen depends on inference depth.
The coda-only configuration $0+2\times13+2$ achieves higher Overall scores at reduced depths, but its Knowledge score declines more strongly during extrapolation.
The balanced configuration $1+2\times13+1$ achieves the highest
Overall and Knowledge scores at the training depth $D=28$.
At $D=56$, however, BaseLoop achieves higher Overall and Reasoning
scores than all three CoreLoop variants.
Thus, the gains from introducing fixed layers depend on both their
placement and the inference budget.

\begin{figure}[t]
    \centering
    \includegraphics[width=\linewidth]{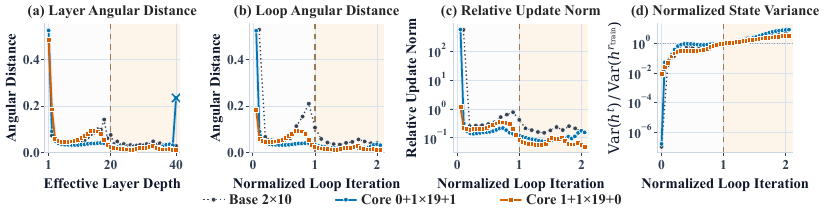}
   \caption{Representation dynamics of BaseLoop and CoreLoop models with two physical layers from Llama3.1-1B.
    The horizontal axes in \textbf{(b)}-\textbf{(d)} are normalized by each configuration's training iteration count.
    \textbf{(a):} The angular distance between consecutive layer states.
    Crosses mark coda layers at the ends of the recorded trajectories.
    \textbf{(b):} The same angular measure between recurrent states
    before and after each complete loop iteration.
    \textbf{(c):} The per-loop update magnitude relative to the
    preceding state, plotted on a logarithmic scale.
    \textbf{(d):} Recurrent-state variance relative to its value at the training iteration count, plotted on a logarithmic scale.
}
    \label{fig:app_baseloop_coreloop_geometry_llama_2}
\end{figure}

\begin{figure}[t]
    \centering
    \setlength{\abovecaptionskip}{5pt}
    \setlength{\belowcaptionskip}{3pt}
    \includegraphics[width=\linewidth]{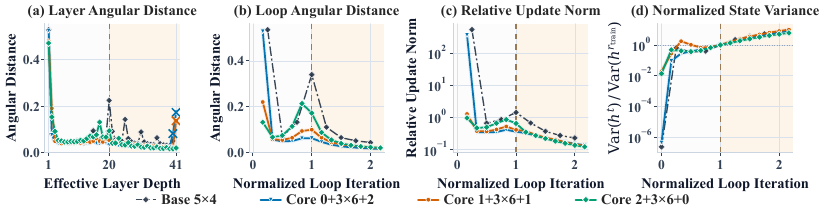}
    \caption{Representation dynamics of BaseLoop and CoreLoop models with five physical layers from Llama3.1-1B.
    BaseLoop $5\times4$ is compared with three CoreLoop variants
    that allocate two fixed layers between the prelude and coda.}
    \label{fig:app_baseloop_coreloop_geometry_llama_5}
\end{figure}

\begin{figure}[t]
    \centering
    \setlength{\abovecaptionskip}{5pt}
    \setlength{\belowcaptionskip}{3pt}
    \includegraphics[width=\linewidth]{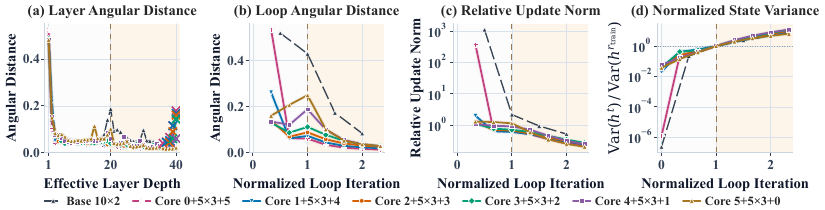}
    \caption{Representation dynamics of BaseLoop and CoreLoop models with ten physical layers from Llama3.1-1B.
    BaseLoop $10\times2$ is compared with six CoreLoop variants
    $p+5\times3+(5-p)$, $p=0,\ldots,5$,
    covering all allocations of five fixed layers between
    the prelude and coda.}
    \label{fig:app_baseloop_coreloop_geometry_llama_10}
\end{figure}

Figs~\ref{fig:app_baseloop_coreloop_geometry_llama_2},
\ref{fig:app_baseloop_coreloop_geometry_llama_5},
and~\ref{fig:app_baseloop_coreloop_geometry_llama_10}
extend the Llama geometry analysis to two, five, and ten physical layers.
At the training loop count, all displayed CoreLoop variants exhibit smaller loop angular distances and relative update norms than their BaseLoop counterparts.
During extrapolation, small angular changes coexist with continued
growth in normalized state variance.
The allocation of fixed layers also changes the trajectories.
For example, with two physical layers, the prelude-only configuration $1+1\times19+0$ ends with smaller recurrent updates and less normalized variance growth than the coda-only configuration $0+1\times19+1$.
For configurations with a coda, the terminal layer-angle increases
correspond to the coda transformations following the recurrent trajectory.

\subsection{Initial-state Input Injection}
\label{app:initial_state_input_injection_results}

\begin{figure}[t]
    \centering
    \includegraphics[width=\linewidth]{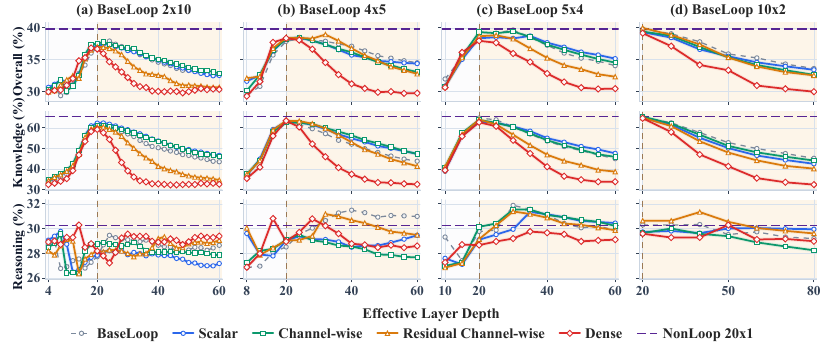}
    \caption{Initial-state input injection on Llama3.1-1B.
        Columns correspond to four shared-stack configurations;
        rows report Overall, Knowledge, and Reasoning scores.
        Curves include all available evaluation depths.
        Vertical dashed lines mark the training depth $D=20$,
        and shading denotes extrapolation.
        BaseLoop uses no injection; NonLoop $20\times1$ is the
        unshared reference.
    }
    \label{fig:app_initial_state_injection_llama}
\end{figure}

The Llama3.1-1B experiments in Fig.~\ref{fig:app_initial_state_injection_llama}
exhibit the similar qualitative pattern as the Qwen results in the main content: initial-state injection provides localized improvements, but does not consistently mitigate degradation beyond the training depth.
For the $4\times5$ configuration, all four injection variants improve Overall at $D=20$, from 37.87 to between 38.21 and 38.42,
yet all fall below BaseLoop at $D=40$
(Tab.~\ref{tab:app_initial_state_injection_llama}).
Dense injection exhibits pronounced Knowledge degradation across all four configurations, while the simpler parameterizations also fail to consistently prevent this decline.
The effects are metric-dependent: for $4\times5$ at $D=40$,
Scalar and Channel-wise injection retain higher Knowledge scores than BaseLoop, but achieve lower Reasoning and Overall scores.
Tabs.~\ref{tab:app_initial_state_injection_llama}
and~\ref{tab:app_initial_state_injection_qwen}
report results for all evaluated parameterizations at the training depth and twice that depth.

\begin{table}[t]
\centering
\scriptsize
\renewcommand{\arraystretch}{1.22}
\setlength{\tabcolsep}{7pt}
\setlength{\abovecaptionskip}{3pt}
\setlength{\belowcaptionskip}{5pt}
\resizebox{\linewidth}{!}{%
\begin{tabular}{llcccccc}
 & & \multicolumn{3}{c}{$\boldsymbol{D=20}$} & \multicolumn{3}{c}{$\boldsymbol{D=40}$}\\
\cmidrule(lr){3-5} \cmidrule(lr){6-8}
\textbf{Configuration} & \textbf{Method} & \textbf{Overall} & \textbf{Knowledge} & \textbf{Reasoning} & \textbf{Overall} & \textbf{Knowledge} & \textbf{Reasoning}\\
\shline
 $2\times10$ & BaseLoop & 36.83 & 61.10 & 27.73 & 34.58 & 50.37 & 28.66\\
 & Scalar & 37.26 & \textbf{62.33} & 27.85 & 34.90 & \textbf{53.64} & 27.87\\
 & Channel-wise & \textbf{37.54} & 60.98 & \textbf{28.75} & \textbf{35.01} & 53.41 & 28.10\\
 & Residual Channel-wise & 37.02 & 60.46 & 28.23 & 32.52 & 41.39 & \textbf{29.19}\\
 & Dense & 36.66 & 59.23 & 28.19 & 29.96 & 32.78 & 28.90\\
\hline
 $4\times5$ & BaseLoop & 37.87 & 62.80 & 28.52 & \textbf{37.11} & 52.11 & \textbf{31.49}\\
 & Scalar & 38.34 & 63.13 & 29.05 & 35.70 & 54.90 & 28.50\\
 & Channel-wise & 38.21 & 62.53 & \textbf{29.09} & 36.15 & \textbf{56.39} & 28.56\\
 & Residual Channel-wise & \textbf{38.42} & \textbf{63.46} & 29.03 & 36.77 & 53.01 & 30.68\\
 & Dense & 38.36 & 63.43 & 28.96 & 31.16 & 37.51 & 28.78\\
\hline
 $5\times4$ & BaseLoop & 39.32 & \textbf{64.79} & 29.77 & 37.26 & 53.64 & 31.12\\
 & Scalar & 38.36 & 63.06 & 29.10 & \textbf{37.67} & \textbf{55.03} & \textbf{31.16}\\
 & Channel-wise & \textbf{39.34} & 63.88 & \textbf{30.14} & 37.35 & 53.99 & 31.10\\
 & Residual Channel-wise & 38.67 & 64.06 & 29.15 & 35.26 & 46.96 & 30.88\\
 & Dense & 37.96 & 62.71 & 28.67 & 32.79 & 41.12 & 29.67\\
\hline
 $10\times2$ & BaseLoop & 39.82 & 65.21 & 30.30 & \textbf{37.80} & \textbf{57.73} & 30.33\\
 & Scalar & 39.33 & 64.92 & 29.73 & 36.60 & 55.18 & 29.64\\
 & Channel-wise & 39.42 & \textbf{65.48} & 29.64 & 36.97 & 56.68 & 29.58\\
 & Residual Channel-wise & \textbf{40.00} & 64.94 & \textbf{30.64} & 37.38 & 53.48 & \textbf{31.34}\\
 & Dense & 39.17 & 64.69 & 29.59 & 34.14 & 47.08 & 29.29\\
\hline
 $20\times1$ & NonLoop & 39.83 & 65.38 & 30.25 & -- & -- & --\\
\end{tabular}%
}
\caption{\footnotesize Full initial-state injection results on Llama3.1-1B at the training depth $D=20$ and twice that depth $D=40$. Configuration denotes the shared-stack depth and training loop count. Bold marks the best result within each recurrent configuration for each metric and evaluation depth. NonLoop is evaluated only at its native depth.}
\label{tab:app_initial_state_injection_llama}
\vspace{-3mm}
\end{table}

\begin{table}[t]
\centering
\scriptsize
\renewcommand{\arraystretch}{1.22}
\setlength{\tabcolsep}{7pt}
\setlength{\abovecaptionskip}{3pt}
\setlength{\belowcaptionskip}{5pt}
\resizebox{\linewidth}{!}{%
\begin{tabular}{llcccccc}
 & & \multicolumn{3}{c}{$\boldsymbol{D=28}$} & \multicolumn{3}{c}{$\boldsymbol{D=56}$}\\
\cmidrule(lr){3-5} \cmidrule(lr){6-8}
\textbf{Configuration} & \textbf{Method} & \textbf{Overall} & \textbf{Knowledge} & \textbf{Reasoning} & \textbf{Overall} & \textbf{Knowledge} & \textbf{Reasoning}\\
\shline
 $2\times14$ & BaseLoop & \textbf{36.85} & \textbf{58.70} & 28.66 & \textbf{37.31} & \textbf{57.30} & \textbf{29.81}\\
 & Scalar & 36.27 & 58.00 & 28.12 & 35.71 & 56.89 & 27.77\\
 & Channel-wise & 35.87 & 58.01 & 27.57 & 35.59 & 56.56 & 27.73\\
 & Residual Channel-wise & 36.77 & 57.88 & \textbf{28.85} & 32.73 & 42.49 & 29.07\\
 & Dense & 35.18 & 55.95 & 27.39 & 30.30 & 33.15 & 29.24\\
\hline
 $4\times7$ & BaseLoop & 36.80 & \textbf{59.44} & 28.31 & 36.59 & \textbf{57.33} & 28.81\\
 & Scalar & 36.16 & 58.35 & 27.84 & 36.34 & 56.05 & 28.95\\
 & Channel-wise & 36.33 & 59.39 & 27.68 & \textbf{36.67} & 57.15 & 29.00\\
 & Residual Channel-wise & \textbf{36.96} & 58.74 & \textbf{28.79} & 36.09 & 51.39 & \textbf{30.35}\\
 & Dense & 36.85 & 58.77 & 28.63 & 31.42 & 36.35 & 29.57\\
\hline
 $7\times4$ & BaseLoop & 37.29 & 61.26 & 28.31 & 36.59 & \textbf{60.33} & 27.68\\
 & Scalar & 36.98 & \textbf{61.90} & 27.64 & 36.64 & 59.55 & 28.04\\
 & Channel-wise & \textbf{37.94} & 61.46 & \textbf{29.12} & \textbf{37.05} & 59.05 & 28.81\\
 & Residual Channel-wise & 37.10 & 61.22 & 28.05 & 36.19 & 58.30 & 27.89\\
 & Dense & 37.04 & 60.74 & 28.15 & 34.19 & 46.49 & \textbf{29.58}\\
\hline
 $14\times2$ & BaseLoop & 37.99 & 61.93 & 29.02 & \textbf{37.87} & 59.95 & 29.59\\
 & Scalar & 38.04 & 61.81 & \textbf{29.12} & 37.23 & 58.41 & 29.29\\
 & Channel-wise & 38.05 & 62.17 & 29.00 & 37.43 & 59.24 & 29.24\\
 & Residual Channel-wise & \textbf{38.27} & \textbf{63.63} & 28.76 & 37.37 & \textbf{60.11} & 28.83\\
 & Dense & 37.75 & 62.52 & 28.46 & 35.66 & 51.78 & \textbf{29.61}\\
\hline
 $28\times1$ & NonLoop & 39.80 & 64.89 & 30.39 & -- & -- & --\\
\end{tabular}%
}
\caption{\footnotesize Full initial-state injection results on Qwen3-0.6B at the training depth $D=28$ and twice that depth $D=56$.}
\label{tab:app_initial_state_injection_qwen}
\vspace{-3mm}
\end{table}

\subsection{History-state Input Injection}
\label{app:history_state_input_injection_results}

Tab.~\ref{tab:app_history_state_input_injection} complements
the history-state injection curves in the main content with numerical results on Qwen3-0.6B under the $4\times7$ shared-stack configuration.
We compare BaseLoop, initial-state injection, and history-state
injection with windows $w\in\{1,2,4\}$ at effective depths
$D=28,56,84$, corresponding to the training depth and
$2\times$ and $3\times$ depth extrapolation.

The preferred history window depends on the parameterization.
At $D=84$, Scalar history injection with $w=4$ achieves the
highest Overall and Knowledge scores among the evaluated
configurations, exceeding BaseLoop by 1.64 and 6.89
percentage points, respectively.
Its Knowledge score remains close to its training-depth value
(58.90 at $D=28$ vs. 59.18 at $D=84$), although its
Reasoning score remains below BaseLoop (29.01 vs. 29.34).
For Channel-wise injection, $w=2$ gives the highest Overall
score among the history variants at all three reported depths,
but does not outperform initial-state injection at $D=84$
(36.14 vs. 36.20).

Dense history injection exhibits a different window preference.
With $w=1$, it achieves the highest Overall score at $D=56$
(37.36) and the highest Reasoning score at $D=84$ (30.77),
while its Knowledge score decreases from 59.72 to 54.16.
Larger windows substantially weaken depth extrapolation:
Dense $w=4$ has the highest Overall score at the training
depth (37.09), but falls to 30.38 at $D=84$.
These results show that training-depth performance does not
reliably predict extrapolation performance, and that improvements
in Overall can reflect different Knowledge--Reasoning trade-offs.

\begin{table*}[t]
\centering
\scriptsize
\renewcommand{\arraystretch}{1.22}
\setlength{\tabcolsep}{7pt}
\setlength{\abovecaptionskip}{3pt}
\setlength{\belowcaptionskip}{5pt}
\resizebox{\linewidth}{!}{%
\begin{tabular}{llccccccccc}
 & & \multicolumn{3}{c}{$\boldsymbol{D=28}$} & \multicolumn{3}{c}{$\boldsymbol{D=56}$} & \multicolumn{3}{c}{$\boldsymbol{D=84}$}\\
\cmidrule(lr){3-5} \cmidrule(lr){6-8} \cmidrule(lr){9-11}
\textbf{Parameterization} & \textbf{Injection} & \textbf{Overall} & \textbf{Knowledge} & \textbf{Reasoning} & \textbf{Overall} & \textbf{Knowledge} & \textbf{Reasoning} & \textbf{Overall} & \textbf{Knowledge} & \textbf{Reasoning}\\
\shline
 -- & BaseLoop & 36.80 & 59.44 & 28.31 & 36.59 & 57.33 & 28.81 & 35.60 & 52.29 & 29.34\\
\hline
 Scalar & Initial-state & 36.16 & 58.35 & 27.84 & 36.34 & 56.05 & 28.95 & 35.69 & 52.10 & 29.53\\
 & History ($w=1$) & 36.45 & 59.41 & 27.84 & 35.74 & 57.67 & 27.51 & 34.69 & 52.23 & 28.12\\
 & History ($w=2$) & 36.46 & 58.83 & 28.07 & 36.78 & 59.18 & 28.38 & 35.75 & 55.83 & 28.22\\
 & History ($w=4$) & 36.61 & 58.90 & 28.25 & 36.80 & \textbf{59.41} & 28.32 & \textbf{37.24} & \textbf{59.18} & 29.01\\
\hline
 Channel-wise & Initial-state & 36.33 & 59.39 & 27.68 & 36.67 & 57.15 & 29.00 & 36.20 & 52.16 & 30.22\\
 & History ($w=1$) & 36.44 & 59.57 & 27.77 & 36.30 & 57.98 & 28.18 & 35.27 & 53.57 & 28.41\\
 & History ($w=2$) & 36.64 & 59.47 & 28.08 & 36.77 & 58.90 & 28.48 & 36.14 & 53.86 & 29.49\\
 & History ($w=4$) & 35.99 & 58.54 & 27.53 & 35.94 & 57.35 & 27.92 & 34.79 & 53.20 & 27.88\\
\hline
 Dense & Initial-state & 36.85 & 58.77 & 28.63 & 31.42 & 36.35 & 29.57 & 29.88 & 32.28 & 28.98\\
 & History ($w=1$) & 36.92 & \textbf{59.72} & 28.38 & \textbf{37.36} & 59.22 & 29.17 & 37.15 & 54.16 & \textbf{30.77}\\
 & History ($w=2$) & 35.96 & 58.84 & 27.38 & 35.99 & 51.96 & \textbf{30.00} & 31.90 & 38.99 & 29.24\\
 & History ($w=4$) & \textbf{37.09} & 58.17 & \textbf{29.18} & 30.54 & 34.23 & 29.16 & 30.38 & 32.73 & 29.50\\
\end{tabular}%
}
\caption{\footnotesize Full history-state input injection results on Qwen3-0.6B with the $4\times7$ shared-stack configuration. $D=28$ is the training depth; $D=56$ and $D=84$ correspond to $2\times$ and $3\times$ depth extrapolation. Initial-state and History denote initial-state-only and history-state-only injection, respectively; $w$ denotes the history window size. BaseLoop uses no injection. All scores are percentages, with higher values indicating better performance. \textbf{Bold} marks the best result in each column across all listed configurations.}
\label{tab:app_history_state_input_injection}
\vspace{-3mm}
\end{table*}

\subsection{Timestep Conditioning}
\label{app:timestep_conditioning_results}

\begin{table}[t]
\centering
\scriptsize
\renewcommand{\arraystretch}{1.22}
\setlength{\tabcolsep}{7pt}
\setlength{\abovecaptionskip}{3pt}
\setlength{\belowcaptionskip}{5pt}
\resizebox{\linewidth}{!}{%
\begin{tabular}{llcccccc}
 & & \multicolumn{3}{c}{$\boldsymbol{D=20}$} & \multicolumn{3}{c}{$\boldsymbol{D=40}$}\\
\cmidrule(lr){3-5} \cmidrule(lr){6-8}
\textbf{Configuration} & \textbf{Method} & \textbf{Overall} & \textbf{Knowledge} & \textbf{Reasoning} & \textbf{Overall} & \textbf{Knowledge} & \textbf{Reasoning}\\
\shline
 $4\times5$ & BaseLoop & 37.87 & 62.80 & 28.52 & 37.11 & 52.11 & \textbf{31.49}\\
 & Loop Gating & 37.99 & \textbf{63.32} & 28.49 & 37.43 & 56.17 & 30.41\\
 & Branch Gating & 37.99 & 62.39 & 28.84 & 37.23 & 56.60 & 29.97\\
 & AdaLN & \textbf{38.58} & 62.65 & \textbf{29.56} & \textbf{37.91} & \textbf{59.68} & 29.75\\
\hline
 $20\times1$ & NonLoop & 39.83 & 65.38 & 30.25 & -- & -- & --\\
\end{tabular}%
}
\caption{\footnotesize Timestep conditioning results on Llama3.1-1B at the training depth $D=20$ and twice that depth $D=40$. All timestep variants use rescaled time grids at inference. Configuration denotes the shared-stack depth and training loop count. Bold marks the best result within the recurrent configuration for each metric and evaluation depth. NonLoop is evaluated only at its native depth.}
\label{tab:app_timestep_conditioning_llama}
\vspace{-3mm}
\end{table}

\begin{table}[t]
\centering
\scriptsize
\renewcommand{\arraystretch}{1.22}
\setlength{\tabcolsep}{7pt}
\setlength{\abovecaptionskip}{3pt}
\setlength{\belowcaptionskip}{5pt}
\resizebox{\linewidth}{!}{%
\begin{tabular}{llcccccc}
 & & \multicolumn{3}{c}{$\boldsymbol{D=28}$} & \multicolumn{3}{c}{$\boldsymbol{D=56}$}\\
\cmidrule(lr){3-5} \cmidrule(lr){6-8}
\textbf{Configuration} & \textbf{Method} & \textbf{Overall} & \textbf{Knowledge} & \textbf{Reasoning} & \textbf{Overall} & \textbf{Knowledge} & \textbf{Reasoning}\\
\shline
 $2\times14$ & BaseLoop & \textbf{36.85} & \textbf{58.70} & \textbf{28.66} & \textbf{37.31} & 57.30 & \textbf{29.81}\\
 & Loop Gating & 36.16 & 57.51 & 28.16 & 35.18 & 57.07 & 26.97\\
 & Branch Gating & 36.62 & 58.20 & 28.53 & 36.64 & \textbf{58.01} & 28.63\\
 & AdaLN & 35.97 & 58.44 & 27.54 & 35.35 & 56.54 & 27.41\\
\hline
 $4\times7$ & BaseLoop & 36.80 & 59.44 & 28.31 & 36.59 & 57.33 & \textbf{28.81}\\
 & Loop Gating & 36.68 & 59.66 & 28.06 & 36.43 & 57.71 & 28.46\\
 & Branch Gating & \textbf{37.16} & \textbf{59.98} & \textbf{28.61} & 36.62 & \textbf{58.60} & 28.38\\
 & AdaLN & 36.18 & 58.48 & 27.81 & \textbf{36.83} & 58.59 & 28.67\\
\hline
 $7\times4$ & BaseLoop & 37.29 & 61.26 & 28.31 & 36.59 & 60.33 & 27.68\\
 & Loop Gating & 37.12 & 60.85 & 28.23 & 36.91 & 59.24 & 28.54\\
 & Branch Gating & 37.27 & 60.73 & 28.48 & \textbf{38.46} & 58.67 & \textbf{30.88}\\
 & AdaLN & \textbf{37.73} & \textbf{61.50} & \textbf{28.81} & 38.01 & \textbf{61.72} & 29.12\\
\hline
 $14\times2$ & BaseLoop & 37.99 & 61.93 & 29.02 & 37.87 & 59.95 & \textbf{29.59}\\
 & Loop Gating & 38.31 & \textbf{63.27} & 28.95 & 37.58 & 59.85 & 29.22\\
 & Branch Gating & \textbf{38.44} & 62.52 & \textbf{29.41} & 37.38 & 59.47 & 29.10\\
 & AdaLN & 38.37 & 63.25 & 29.04 & \textbf{38.05} & \textbf{61.24} & 29.36\\
\hline
 $28\times1$ & NonLoop & 39.80 & 64.89 & 30.39 & -- & -- & --\\
\end{tabular}%
}
\caption{\footnotesize Timestep conditioning results on Qwen3-0.6B at the training depth $D=28$ and twice that depth $D=56$. All timestep variants use rescaled time grids at inference. Configuration denotes the shared-stack depth and training loop count. Bold marks the best result within each recurrent configuration for each metric and evaluation depth. NonLoop is evaluated only at its native depth.}
\label{tab:app_timestep_conditioning_qwen}
\vspace{-3mm}
\end{table}

Tabs~\ref{tab:app_timestep_conditioning_llama}
and~\ref{tab:app_timestep_conditioning_qwen} show that the benefits of timestep conditioning depend on the recurrent configuration and task category.
On Qwen3-0.6B, Branch Gating provides the largest Overall gain
at twice the training depth for $7\times4$
(38.46 vs.\ 36.59 for BaseLoop), driven by improved Reasoning,
whereas AdaLN achieves the highest Knowledge score (61.72).
However, no conditioning variant improves Overall for $2\times14$
at either reported depth, and the training-depth gains of both
scalar gating schemes for $14\times2$ disappear at $D=56$.
On Llama3.1-1B, all three variants improve Overall and Knowledge
at $D=40$, with AdaLN performing best on both metrics, but none
matches BaseLoop in Reasoning.
Thus, timestep conditioning can improve performance beyond the
training depth, but no variant consistently dominates across
configurations and metrics.

\subsection{Combination Experiments}
\label{app:combine_results}

\paragraph{Initial-state and history-state.}

\begin{figure}[tb]
    \centering
    \setlength{\abovecaptionskip}{4pt}
    \setlength{\belowcaptionskip}{5pt}
    \includegraphics[width=1.0\linewidth]{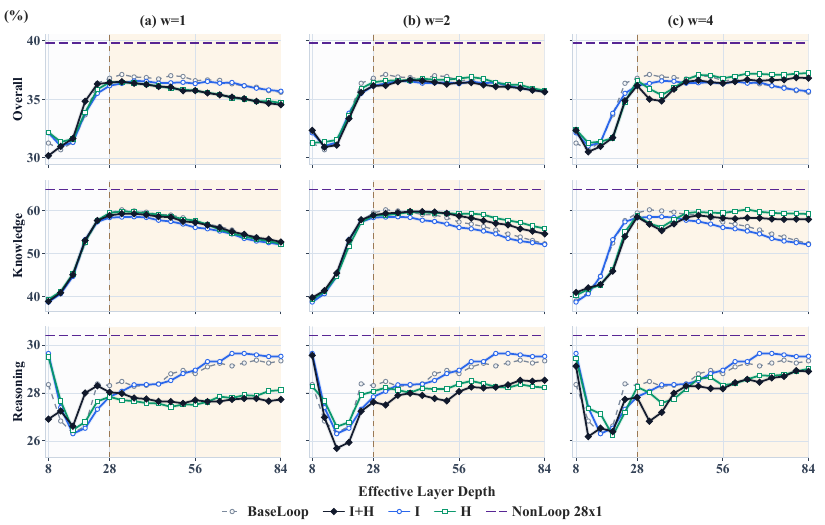}
    \caption{Results of Scalar initial plus history injection combination and their individual components across history windows $w=1,2,4$ with the Qwen3-0.6B $4\times7$ configuration.}
    \label{fig:initial_history_scalar}
\end{figure}

\begin{figure}[tb]
    \centering
    \setlength{\abovecaptionskip}{4pt}
    \setlength{\belowcaptionskip}{5pt}
    \includegraphics[width=1.0\linewidth]{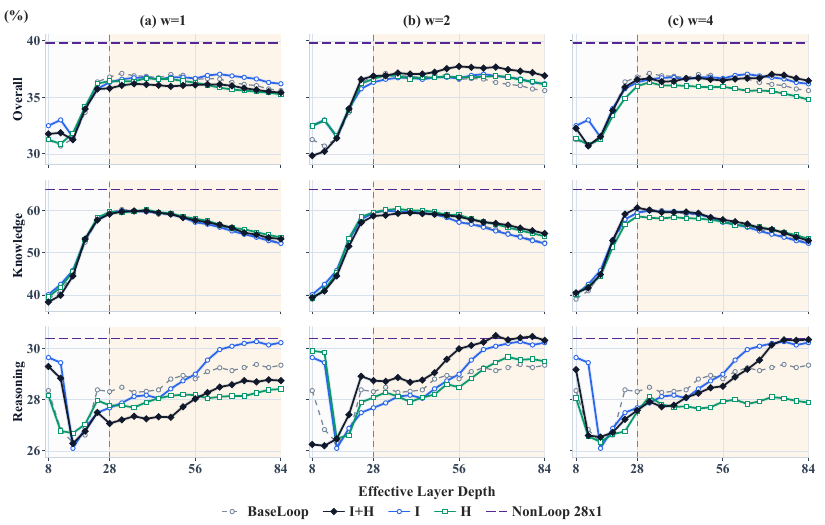}
    \caption{Results of Channel-wise initial plus history injection combination and their individual components across history windows $w=1,2,4$ with the Qwen3-0.6B $4\times7$ configuration.}
    \label{fig:initial_history_vector}
\end{figure}

\begin{figure}[tb]
    \centering
    \setlength{\abovecaptionskip}{4pt}
    \setlength{\belowcaptionskip}{5pt}
    \includegraphics[width=1.0\linewidth]{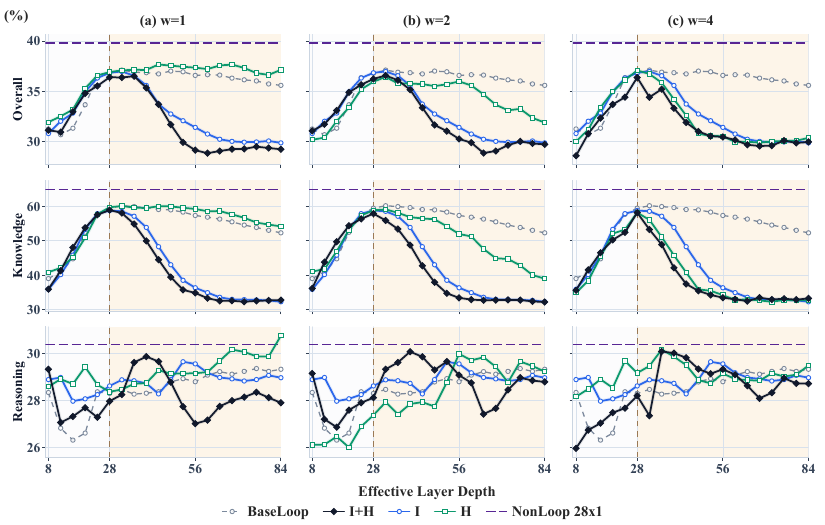}
    \caption{Results of Dense initial plus history injection combination and their individual components across history windows $w=1,2,4$ with the Qwen3-0.6B $4\times7$ configuration.}
    \label{fig:initial_history_linear}
\end{figure}

Full results of Initial-state injection plus history-state injection are provided in Figs.~\ref{fig:initial_history_scalar},~\ref{fig:initial_history_vector}, and~\ref{fig:initial_history_linear}.
At the training depth of 28, most combined runs have
overall scores near 37. 
Beyond this depth, the scalar and channel-wise
combinations remain substantially more stable than the dense combinations.
For scalar injection with $w=4$, combined knowledge reaches 57.92 at depth 84, compared with 52.10 for input injection alone, although history
injection alone reaches 59.18. 
The clearest gain over both components occurs for channel-wise injection with $w=2$: at depth 56, its overall score
is 37.74, vs. 36.67 and 36.77 for initial-state and history-state injection alone.
Dense combinations instead lose substantial overall and knowledge accuracy beyond depth 28. 
For $w=1$ at depth 84, combined knowledge falls to 32.77,
while history injection alone retains 54.16. 

\begin{figure}[tb]
    \centering
    \setlength{\abovecaptionskip}{4pt}
    \setlength{\belowcaptionskip}{5pt}
    \includegraphics[width=1.0\linewidth]{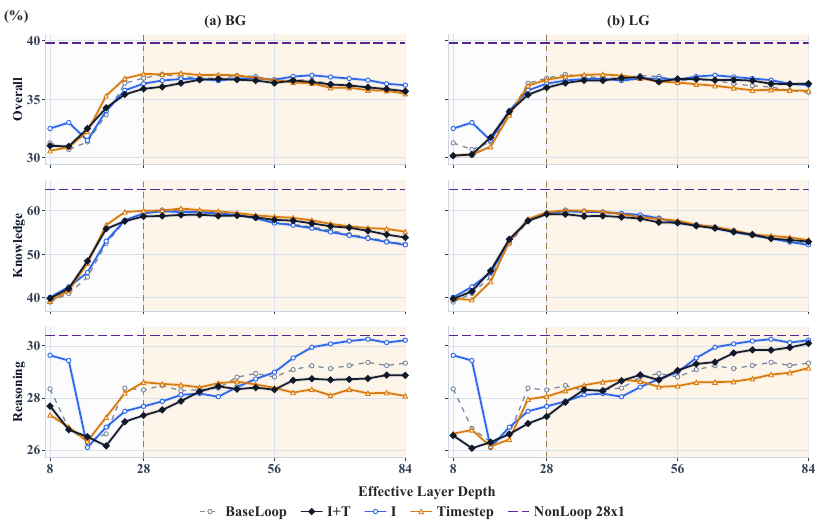}
    \caption{Results of Dense initial plus timestep combination and their individual components with the Qwen3-0.6B $4\times7$ configuration.}
    \label{fig:initial_timestep}
\end{figure}

\paragraph{Initial-state and timestep.}
Fig.~\ref{fig:initial_timestep} compares initial-state injection combined with timestep conditioning against the individual components. 
Combining the two provides almost no consistent benefit under either BG or LG. 
At the training depth $D=28$, the combined Overall
scores are 35.89 with BG and 35.99 with LG, below
the corresponding gating-only scores of 37.16
and 36.68, respectively.
At depth 56, LG exceeds initial-state injection by only 0.07 percentage points, while BG remains below both components. 
This pattern persists at depth 84: LG gains just
0.13 points over initial-state injection, whereas BG trails it by 0.52 points. 
Knowledge and reasoning likewise show no reliable joint gain.
Thus, timestep conditioning does not provide a meaningful complementary improvement when added to initial-state injection in these runs.

\begin{figure}[h]
    \centering
    \setlength{\abovecaptionskip}{4pt}
    \setlength{\belowcaptionskip}{5pt}
    \includegraphics[width=1.0\linewidth]{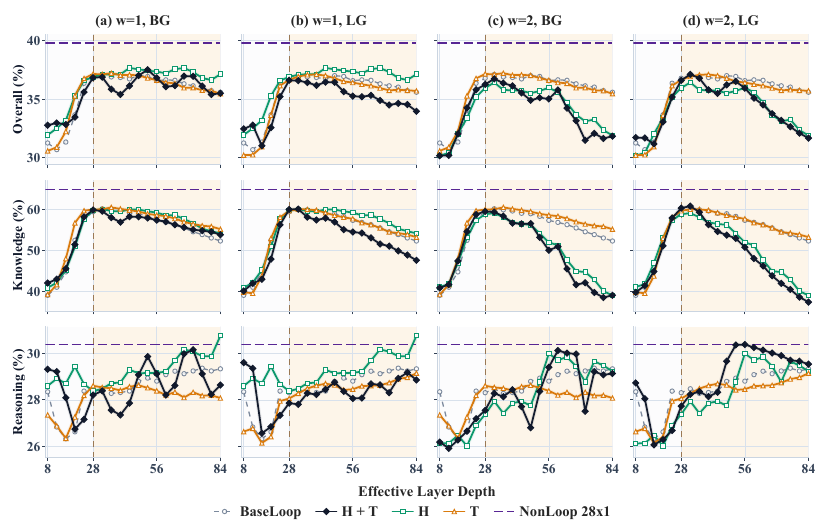}
    \caption{Results of Dense history injection plus timestep combination and their individual components for history windows $w=1,2$ with the Qwen3-0.6B $4\times7$ configuration.}
    \label{fig:history_timestep_1}
\end{figure}

\begin{figure}[h]
    \centering
    \setlength{\abovecaptionskip}{4pt}
    \setlength{\belowcaptionskip}{5pt}
    \includegraphics[width=1.0\linewidth]{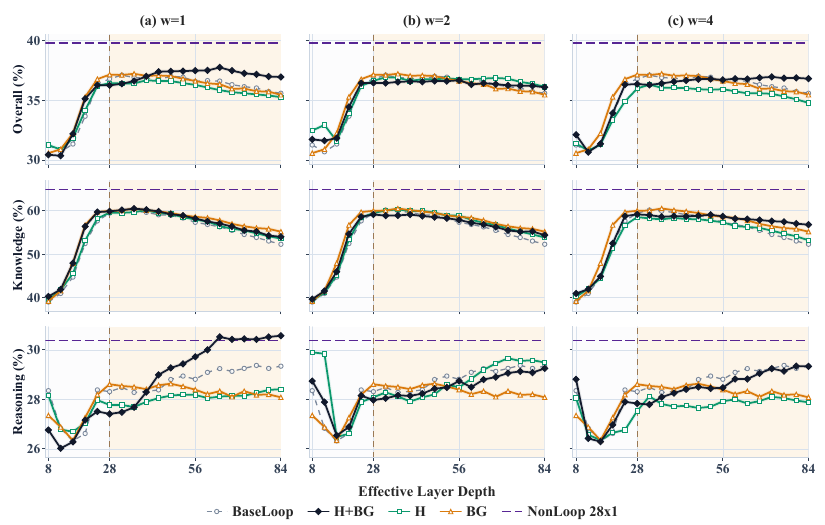}
    \caption{Results of Channel-wise history injection combined with BG and their individual components across history windows $w=1,2,4$ with the Qwen3-0.6B $4\times7$ configuration.}
    \label{fig:history_timestep_2}
\end{figure}

\begin{figure}[h]
    \centering
    \setlength{\abovecaptionskip}{4pt}
    \setlength{\belowcaptionskip}{5pt}
    \includegraphics[width=1.0\linewidth]{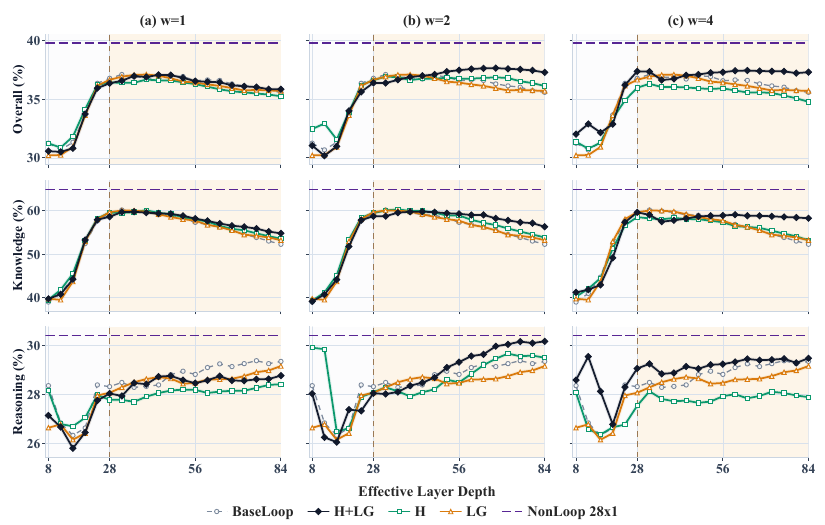}
    \caption{Results of Channel-wise history injection combined with LG and their individual components across history windows $w=1,2,4$ with the Qwen3-0.6B $4\times7$ configuration.}
    \label{fig:history_timestep_3}
\end{figure}

\paragraph{History-state and timestep.}
Figs~\ref{fig:history_timestep_1}, \ref{fig:history_timestep_2}, and
\ref{fig:history_timestep_3} compare history-state injection with timestep conditioning. 
It is clear that several history-state combinations show clear gains at greater effective depths.
At depth 84, channel-wise history with $w=1$ and BG scores 36.96 overall, compared with 35.68 for initial-state plus BG. 
With $w=2$ and LG, the corresponding scores are 37.30 and 36.33; the history-state combination also improves knowledge from 52.92 to 56.32. 
A channel-wise history signal is already effective on its own: at depth 84, history-only $w=2$ scores 36.14 overall, versus 31.90 for dense history-only $w=2$. 
Adding LG raises the channel-wise result to 37.30, while dense history with the same window and gate scores
31.68. 
Thus, a dense history transformation is unnecessary for the
strongest results here. 
Channel-wise history uses $wd$ injection weights,
compared with $wd^2$ for dense history; at $d=1024$ and $w=2$, this is 2{,}048 vs. 2{,}097{,}152 history-injection weights.

\paragraph{Initial-state, history-state, and timestep.}

\begin{figure}[h]
    \centering
    \setlength{\abovecaptionskip}{5pt}
    \setlength{\belowcaptionskip}{-2pt}
    \includegraphics[width=1.0\linewidth]{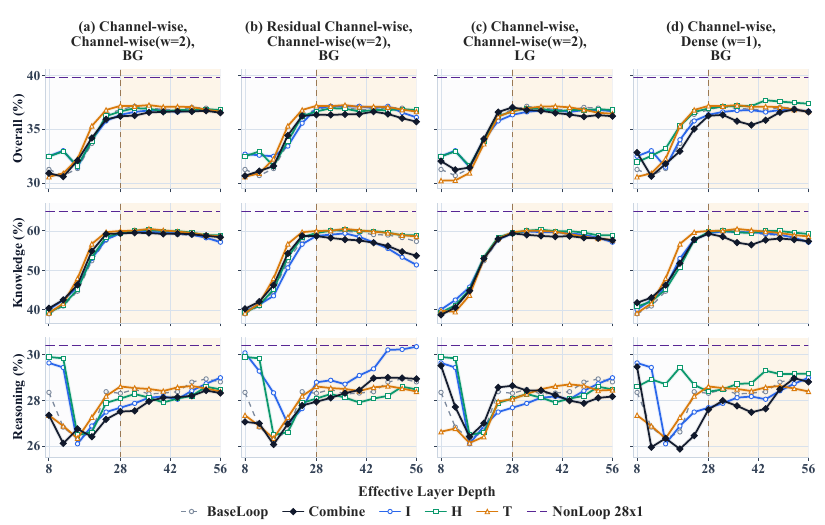}
    \caption{Results of three-mechanism combination and their individual components across history windows $w=1,2$ with the Qwen3-0.6B $4\times7$ configuration.}
    \label{fig:combination_all}
\end{figure}

Finally, Fig.~\ref{fig:combination_all} evaluates all three mechanisms together. 
Within the evaluated range, adding a third mechanism does not
produce a consistent additive gain. 
The channel-wise initial-state and history-state combination with LG reaches an Overall score of 37.02 at the training depth $D=28$, exceeding the best individual component, LG (36.68), by 0.34 percentage points.
However, this advantage does not persist at greater depths.
At $D=56$, the four three-mechanism configurations achieve Overall scores of 35.69-36.60; none exceeds its best matched individual component.
Their scores are also below the reported two-mechanism results of 37.74 for initial-state plus history-state injection with $w=2$, and 37.49 for channel-wise history-state injection plus LG with $w=2$.
Knowledge and reasoning show no consistent compensating gain. 
Among the evaluated configurations, combining all three mechanisms does not consistently outperform the two-mechanism alternatives.

\section{Further Analysis}

\subsection{Input Noise Initialization Ablation}

\begin{figure}[h]
    \centering
    \setlength{\abovecaptionskip}{2pt}
    \setlength{\belowcaptionskip}{-2pt}
    \includegraphics[width=1.0\linewidth]{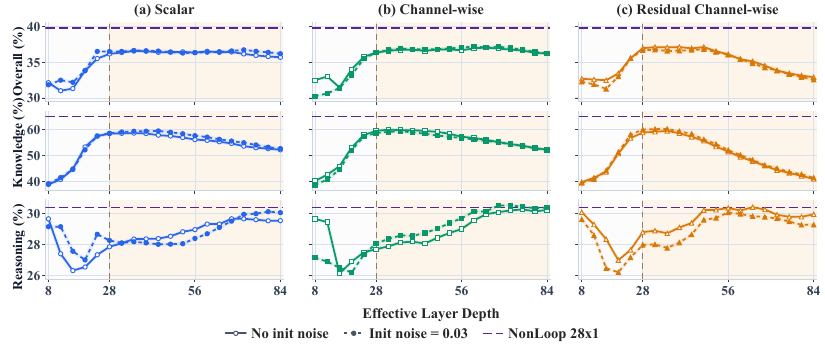}
    \caption{Average results for Qwen3-0.6B BaseLoop $4\times7$ with Scalar, Channel-wise, and Residual Channel-wise initial-state input injection under an extended effective layer of 84 (3x training budget). The vertical boundary marks the training depth of 28 layers, with shading indicating depth extrapolation. Initialization noise has a limited effect on performance, supporting our default choice of initialization without noise.}
    \label{fig:noise_ablation_results}
\end{figure}

We study the effect of initialization noise in initial-state injection using Qwen3-0.6B with a BaseLoop $4\times7$ configuration, comprising four shared layers trained for seven loops. 
We compare initialization without noise against a noise scale of 0.03 for Scalar, Channel-wise, and Residual Channel-wise input injection, evaluating overall, knowledge, and reasoning performance across inference loops. 
As shown in Fig.~\ref{fig:noise_ablation_results}, adding noise has only a minor effect on downstream performance and both initialization settings exhibit similar trends across evaluation depths during an extended inference of 84 effective layers. 
The small differences do not indicate a consistent advantage from noise across variants and metrics. 
We therefore adopt initialization without noise as the default choice.

\subsection{Geometry Analysis}

\begin{figure}[t]
    \centering
    \setlength{\abovecaptionskip}{2pt}
    \setlength{\belowcaptionskip}{-3pt}
    \includegraphics[width=1.0\linewidth]{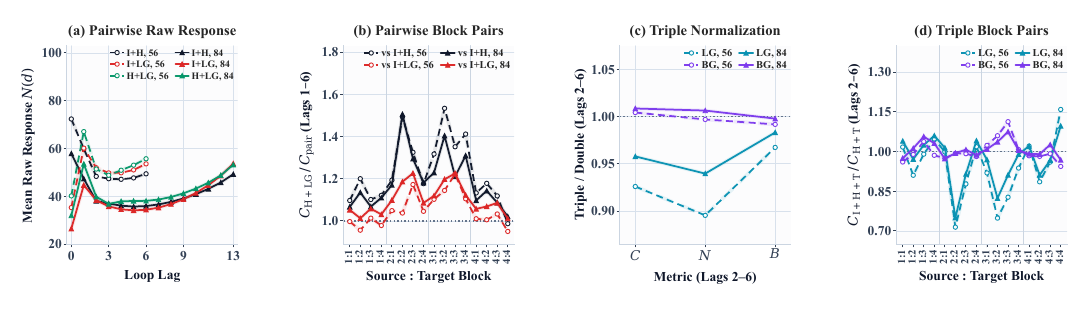}
    \caption{\footnotesize Additional computational interaction results for Qwen3-0.6B BaseLoop $4\times7$, using the probes defined in App.~\ref{app:probing}. All state conditioning is channel-wise, with history window $w=2$.
    \textbf{(a)} Mean unnormalized response $N(d)$ for pairwise conditioning, showing all measured lags.
    \textbf{(b)} Ratios of mean $C$ for H+LG versus I+H and I+LG, separately for each of the 16 shared source/target block pairs over lags $1$-$6$.
    \textbf{(c)} Triple/double ratios of mean $C$, mean $N$, and mean $B$
    over lags $2$-$6$, with the gate type and history settings matched.
    \textbf{(d)} Triple/double ratios of mean $C$ for each shared-block
    pair over lags $2$-$6$.
    All means weight occurrence pairs equally on the indicated lag support;
    $B_j$ is repeated for each included source paired with target $j$.
    Configuration ratios divide these means, and mean $C$ is the mean of
    $N_{i\to j}/B_j$, not the ratio of mean $N$ to mean $B$.
    In (\textbf{b,d}), $s{:}q$ denotes one-based source and target block indices;
    vertical separators distinguish source blocks. Lines between block
    pairs or metrics are visual guides.
    Dashed curves with open circles denote $D=56$; solid curves with filled
    triangles denote $D=84$. Horizontal references at one indicate equal
    values.}
    \label{fig:appendix_composition_geometry}
\end{figure}

Complementing the geometry analysis in
Fig.~\ref{fig:conditioning_responses}\textbf{(c,d)},
Fig.~\ref{fig:appendix_composition_geometry} examines unnormalized update
responses and variation across shared-block pairings. 
At $D=84$, H+LG has $6.4\%$ and $10.2\%$ higher mean raw response over lags $1$-$6$ than I+H and I+LG, respectively, while its mean relative response is higher in all 16 shared-block pairings for both comparisons
(\textbf{a,b}). 
Thus, the cross-loop interaction advantage identified in the main text also appears in the response numerator and extends across shared-block pairings. 
Adding initial-state conditioning to H+LG
reduces the mean raw response over lags $2$-$6$ by $10.5\%$ at $D=56$
and $6.0\%$ at $D=84$, supporting the main text's observation that the
lower relative response is not solely a normalization effect
(\textbf{c}). 
This reduction is heterogeneous: at $D=84$, the LG
triple has lower mean $C$ in $9/16$ shared-block pairings, while the
remaining pairings increase; the BG triple's aggregate responses remain
close to those of H+BG (\textbf{c,d}). 
These results further support the interpretation of history-state and timestep complementarity through cross-loop interaction, and provide a possible explanation for the lack of additive gains from initial-state conditioning, without implying a uniform reduction across gates or block pairings.

\end{document}